\documentclass[10pt,twocolumn,letterpaper]{article}

\usepackage[cvprfinal]{cvpr}
\usepackage[utf8]{inputenc}
\usepackage{times}
\usepackage{epsfig}
\usepackage{graphicx}
\usepackage{float}
\usepackage{amsmath}
\usepackage{pifont}
\usepackage{adjustbox}
\usepackage{amssymb}
\usepackage{multirow}
\usepackage{booktabs}
\usepackage[accsupp]{axessibility}
\usepackage{hhline}
\usepackage[skip=1ex,font=small,labelsep=period]{caption}
\usepackage{enumitem}
\usepackage{tabularx}
\usepackage{mathtools}

\usepackage[pagebackref,breaklinks,colorlinks]{hyperref}

\newcolumntype{Y}{>{\centering\arraybackslash}X}
\newcolumntype{R}{>{\raggedleft\arraybackslash}X}
\newcolumntype{L}{>{\raggedright\arraybackslash}X}

\makeatletter
\@namedef{ver@everyshi.sty}{}
\makeatother
\usepackage{tikz}

\usepackage[capitalize]{cleveref}
\crefname{section}{Sec.}{Secs.}
\Crefname{section}{Section}{Sections}
\Crefname{table}{Table}{Tables}
\crefname{table}{Tab.}{Tabs.}

\newcommand{\Model}{S3F}

\newcommand{\cmark}{\ding{51}}%
\newcommand{\xmark}{\ding{55}}%
\newcolumntype{R}[2]{%
    >{\adjustbox{angle=#1,lap=\width-(#2)}\bgroup}%
    l%
    <{\egroup}%
}
\newcommand*\rot{\multicolumn{1}{R{35}{1em}}}

\renewcommand{\vec}[1]{\boldsymbol{#1}}
\newcommand{\mat}[1]{\mathbf{#1}}

\newcommand{\real}[0]{\mathbb{R}}

\newcommand{\image}[0]{\mat{I}}

\newcommand{\pose}[0]{\vec{\theta}}
\newcommand{\shape}[0]{\vec{\beta}}

\newcommand{\ba}{\boldsymbol{a}}

\newcommand{\bc}{\boldsymbol{c}}

\newcommand{\be}{\boldsymbol{e}}
\newcommand{\bff}{\boldsymbol{f}}

\newcommand{\bn}{\boldsymbol{n}}

\newcommand{\br}{\boldsymbol{r}}
\newcommand{\bs}{\boldsymbol{s}}

\newcommand{\bv}{\boldsymbol{v}}

\newcommand{\bx}{\boldsymbol{x}}

\newcommand{\bF}{\mathbf{F}}

\newcommand{\bI}{\mathbf{I}}

\newcommand{\bL}{\mathbf{L}}

\newcommand{\bO}{\mathbf{O}}

\newcommand{\bV}{\mathbf{V}}

\def\cvprPaperID{4307}
\def\confName{CVPR}
\def\confYear{2023}

\begin{document}

\title{Structured 3D Features for Reconstructing Controllable Avatars}

\author{
Enric Corona
\hspace{2ex} Mihai Zanfir
\hspace{2ex} Thiemo Alldieck \\
\hspace{2ex} Eduard Gabriel Bazavan
\hspace{2ex} Andrei Zanfir
\hspace{2ex} Cristian Sminchisescu 
\vspace{1.3ex}
\\
\textbf{Google Research}
}

\twocolumn[{
\renewcommand\twocolumn[1][]{#1}
\vspace{-0.7cm}
\maketitle
\thispagestyle{empty}
\vspace{-0.55cm}
\includegraphics[width=1\linewidth, trim={0.2cm 0 0.2cm 0}, clip=true]{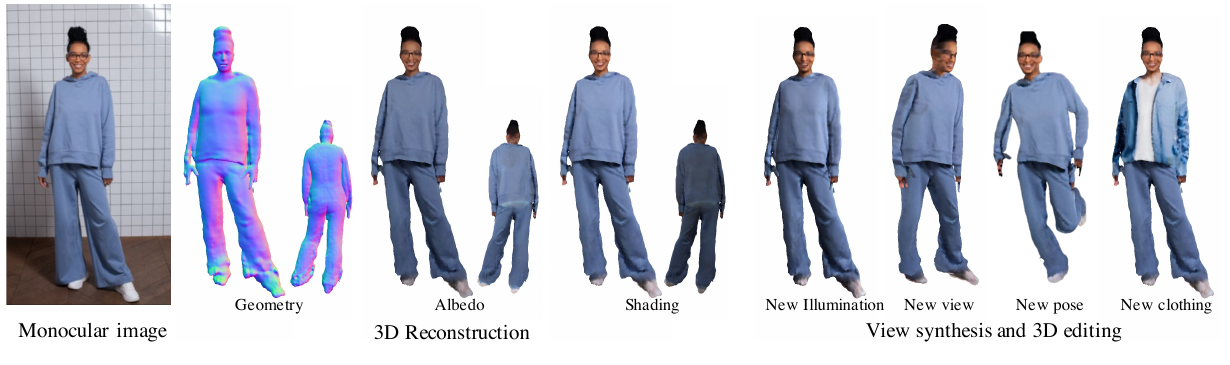} \\
\vspace{-0.75cm}
\captionof{figure}{%
\small{We introduce \textit{Structured 3D Features} (S3F), a new feature representation for monocular 3D Human Reconstruction.
Using S3Fs, we train a new implicit reconstruction model for creating rigged and relightable avatars -- without the need for post-processing.
S3Fs store features in a dense 3D point cloud around a human body model, allowing for various applications: our model can produce 3D reconstructions in novel poses, under new illumination, under novel views, or with altered clothing for virtual try-on. 
}}
\label{fig:teaser}
\vspace{0.59cm}
}]

\begin{abstract}\vspace{-2mm}
We introduce \textit{Structured 3D Features}, a model based on a novel implicit 3D representation that pools pixel-aligned image features onto dense 3D points sampled from a parametric, statistical human mesh surface. The 3D points have associated semantics and can move freely in 3D space. This allows for optimal coverage of the person of interest, beyond just the body shape, which in turn, additionally helps modeling accessories, hair, and loose clothing.
Owing to this, we present a complete 3D transformer-based attention framework which, given a single image of a person in an unconstrained pose, generates an animatable 3D reconstruction with albedo and illumination decomposition, as a result of a single end-to-end model, trained semi-supervised, and with no additional postprocessing. %
We show that our \Model~model surpasses the previous state-of-the-art on various tasks, including monocular 3D reconstruction, as well as albedo \& shading estimation. Moreover, we show that the proposed methodology allows novel view synthesis, relighting, and re-posing the reconstruction, and can naturally be extended to handle multiple input images (\eg different views of a person, or the same view, in different poses, in video). Finally, we demonstrate the editing capabilities of our model for 3D virtual try-on applications. 

\end{abstract}

\vspace{-0.2cm}
\section{Introduction}

Human digitization is playing a major role in several important applications, including AR/VR, video games, social telepresence, virtual try-on, or the movie industry.
Traditionally, 3D virtual avatars have been created using multi-view stereo \cite{li2021topologically} or expensive equipment~\cite{remelli2022drivable}.
More flexible or low-cost solutions are often template-based~\cite{alldieck2019learning,tex2shape}, but these lack expressiveness for representing details such as hair.
Recently, research on implicit representations~\cite{saito2019pifu, saito2020pifuhd,phorhum,huang2020arch,smplicit} and neural fields~\cite{zheng2022avatar,jiang2022selfrecon,xu2021h,neuralbody} has made significant progress in improving the realism of avatars.%
The proposed models produce detailed results and are also capable, within limits, to represent loose hair or clothing.

Even though model-free methods yield high-fidelity avatars, they are not suited for downstream tasks such as animation.
Furthermore, proficiency is often limited to a certain range of imaged body poses.
Aiming at these problems, different efforts have been made to combine parametric models with more flexible implicit representations~\cite{huang2020arch,he2021arch++,xiu2022icon,zheng2021pamir}.
These methods support animation~\cite{huang2020arch, he2021arch++} or tackle challenging body poses~\cite{xiu2022icon, zheng2021pamir}.
However, most work relies on 2D pixel-aligned features.
This leads to two important problems: (1) First, errors in body pose or camera parameter estimation will result in misalignment between the projections of 3D points on the body surface and image features, which ultimately results in low quality reconstructions.
(2) image features are coupled with the input view and cannot be easily manipulated \eg~to edit the reconstructed body pose.

In this paper we introduce Structure 3D Features (\Model), a flexible extension to image features, specifically designed to tackle the previously discussed challenges and to provide more flexibility during and after digitization.
\Model~store local features on ordered sets of points around the body surface, taking advantage of the geometric body prior.
As body models do not usually represent hair or loose clothing, it is difficult to recover accurate body parameters for images in-the-wild.
To this end, instead of relying too much on the geometric body prior, our model freely moves 3D body points independently to cover areas that are not well represented by the prior.
This process results in our novel S3Fs, and is trained without explicit supervision only using reconstruction losses as signals.
Another limitation of prior work is its dependence on 3D scans.
We alleviate this dependence by following a mixed strategy: we combine a small collection of 3D scans, typically the only training data considered in previous work, with large-scale monocular in-the-wild image collections. 
We show that by guiding the training process with a small set of 3D synthetic scans, the method can efficiently learn features that are only available for the scans (\eg. albedo), while self-supervision on real images allow the method to generalize better to diverse appearances, clothing types and challenging body poses.

In this paper, we show how the proposed \textit{S3Fs} are substantially more flexible than current state-of-the-art representations.
S3Fs enable us to train a single end-to-end model that, based on an input image and matching body pose parameters, can generate a 3D human reconstruction that is relightable and animatable.
Furthermore, our model supports the 3D editing of \eg clothing, without additional post-processing.
See Fig.\ref{fig:teaser} for an illustration and Table~\ref{tab:positioning} for a summary of our model's properties, in relation to prior work.
We compare our method with prior work and demonstrate state-of-the-art performance for monocular 3D human reconstruction from challenging, real-world and unconstrained images, and for albedo \& shading estimation.
We also provide an extensive ablation study to validate our different architectural choices and training setup.
Finally, we show how the proposed approach can also be naturally extended to integrate observations across different views or body poses, further increasing reconstruction quality.

\begin{table}
\centering
\setlength{\tabcolsep}{8pt}
\resizebox{0.9\linewidth}{!}{%
    \begin{tabular}{cccccccc|l}
    \rot{end-to-end trainable} & \rot{returns albedo}  & \rot{returns shading} & \rot{true surface normals} & \rot{challenging poses} & \rot{animatable} & \rot{semantic editing} & \rot{allows multi-view} & \\
    \hline
    \cmark & \xmark & \cmark & \cmark & \xmark & \xmark & \xmark & \cmark & PIFu~\cite{saito2019pifu} \\
    \xmark & \xmark & \xmark & \cmark & \xmark & \xmark & \xmark & \xmark & PIFuHD~\cite{saito2020pifuhd} \\
    \cmark & \xmark & \cmark & \xmark & \xmark & \cmark & \xmark & \xmark & ARCH++ \cite{he2021arch++} \\
    \xmark & \xmark & \cmark & \cmark & \cmark & \cmark & \xmark & \xmark & PaMIR~\cite{zheng2021pamir} \\
    \cmark & \cmark & \cmark & \cmark & \xmark & \xmark & \xmark & \xmark & PHORHUM~\cite{phorhum} \\
    \xmark & \xmark & \xmark & \cmark & \cmark & \cmark & \xmark & \xmark & ICON~\cite{xiu2022icon} \\
    \hline
    \cmark &\cmark & \cmark & \cmark & \cmark & \cmark & \cmark & \cmark & \textbf{\Model~(Ours)} \\
\end{tabular}}
\caption{\small{\textbf{Key properties of \Model~ compared to recent work}. Our model includes a number of desirable novel features with respect to previous state-of-the-art, and recovers rigged and relightable human avatars even for challenging body poses in input images.
}}
\label{tab:positioning}
\end{table}
\vspace{-0.1cm}
\section{Related work}

Table~\ref{tab:positioning} summarizes the main properties of the most recent generative cloth models we have discussed.

\vspace{1mm}
\noindent{\bf Monocular 3D Human Reconstruction}
is an inherently ill-posed problem and thus greatly benefits from  strong human body priors.
The reconstructed 3D shape of a human is often a byproduct of 3D pose estimation~\cite{hmr,spin,kolotouros2019convolutional, lassner2017unite,omran2018neural,smplx,frankmocap,smith2019facsimile, zanfir2021thundr,corona2022learned} represented by a statistical human body model~\cite{smpl, smplx, xu2020ghum, mano}. 
Body models, however, only provide mid-resolution body meshes that do not capture important elements of a person's detail, such as clothing or accessories.
To this end, one line of work extends parametric bodies to represent clothing through offsets to the body mesh~\cite{alldieck2019learning, alldieck2018detailed, alldieck2018video,onizuka2020tetratsdf,tex2shape,bhatnagar2019multi,zhu2019detailed}.
However, this is prone to fail for loose garments or those with a different topology than the human body. Other representations of the clothed human body have been explored, including voxels~\cite{varol2018bodynet, zheng2019deephuman}, geometry images~\cite{pumarola20193dpeople}, bi-planar depth maps~\cite{gabeur2019moulding} or visual hulls~\cite{natsume2019siclope}. To date, the most powerful representations are implicit functions~\cite{saito2019pifu,saito2020pifuhd,phorhum, xiu2022icon, huang2020arch,  he2021arch++, zheng2021pamir,integratedpifu,cao22jiff,yang2021s3} that define 3D geometry via a decision boundary~\cite{mescheder2019occupancy,chen2019learning} or level set~\cite{deepsdf}.
A popular choice to condition implicit functions on an input image are pixel-aligned features~\cite{saito2019pifu}.
This approach has been used to obtained detailed 3D reconstruction methods without relying on a body template~\cite{saito2019pifu, saito2019pifu, phorhum,yang2021s3}, and in combination with parametric models~\cite{huang2020arch, he2021arch++, xiu2022icon,yang2021s3} to take advantage of body priors.
ARCH and ARCH++ map pixel-aligned features to a canonical space of SMPL~\cite{smpl} to support animatable reconstructions.
However, this requires almost perfect SMPL estimates, which are hard to obtain for images in-the-wild.
To address this, ICON~\cite{xiu2022icon} proposes an iterative refinement of the SMPL parameters during 3D reconstruction calculations.
ICON requires different modules for predicting front \& back normal maps, to compute features, and for SMPL fitting.
In contrast, we propose an end-to-end trainable model that can correct potential errors in body pose and shape estimation, without supervision, by freely allocating relevant features around the approximate body geometry.

Somewhat related are also methods for human relighting in monocular images \cite{lagunas2021single, ji2022geometry, kanamori2019relighting, tajima2021relighting}.
However, these methods typically do not reconstruct the 3D human and instead rely on normal maps to transform pixel colors.
Our work bears similarity with PHORHUM~\cite{phorhum}, in our prediction of albedo, and a global scene illumination code. However, we combine this approach with mixed supervision from both synthetic and real data, thus obtaining detailed, photo-realistic reconstructions, for images in-the-wild.

\vspace{1mm}
\noindent{\bf Human neural fields}. 
Our work is also related to neural radiance fields, or NeRFs~\cite{nerf}, from which we take inspiration for our losses on images in-the-wild. 
NeRFs have been recently explored for novel human view synthesis~\cite{neuralbody, chen2021animatable, peng2021animatable,su2021nerf,xu2021h, corona2022lisa,arah_eccv22,jiang2022neuman,weng2022humannerf}. %
These methods are trained per subject by minimizing residuals between rendered and observed images and are typically based on a parametric body model. 
See~\cite{tewari2022advances} for an extensive literature review.
In contrast, our method is trained on images of different subjects and thus generalizes to unseen identities. Finally, our work follows LVD~\cite{corona2022learned} by predicting a neural deformation field for body vertices, in our case with no explicit supervision.

\begin{figure*}[t!]
\vspace{-0.42cm}
\begin{center}
\includegraphics[width=.9\linewidth, trim={0 0.2cm 0 0.2cm}, clip=true]{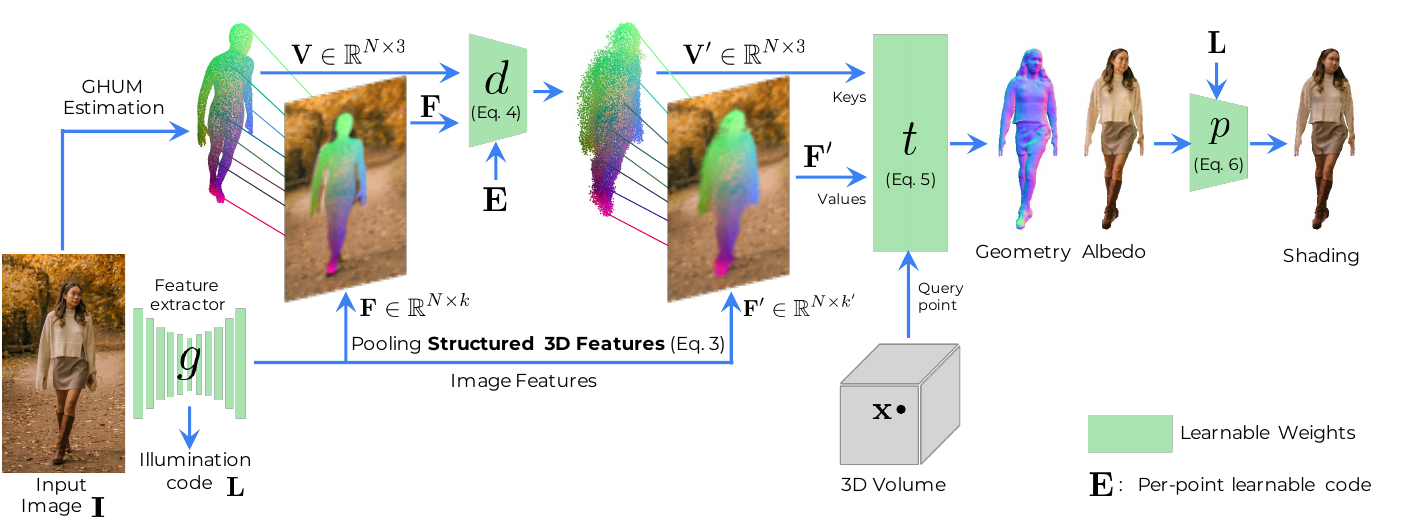}
\vspace{-0.1cm}
\caption{\small{\textbf{Method overview.}
We introduce a new implicit 3D representation S3F (Structured 3D Features) that utilizes $N$ points sampled on the body surface $\bV$ and their 2D projection to pool features $\bF$ from a 2D feature map extracted from the input image $\bI$.
The initial body points are non-rigidly displaced by the network $d$ to obtain $\bV'$ in order to sample new features $\bF'$, on locations that are not covered by body vertices, such as loose clothing or hair.
Given an input point $x$, we then aggregate representations from the set of points and their features using a transformer architecture $t$, to finally obtain per-point signed distance and albedo color. Finally, we can relight the reconstruction using the predicted albedo, an illumination representation $L$ and a shading network $p$. 
We assume perspective projection to obtain more natural reconstructions, with correct proportions.}
\label{fig:diagram}}
\end{center}
\vspace{-0.25cm}
\end{figure*}

\vspace{-0.1cm}
\section{Method}\label{sec:approach}

We seek to estimate the textured and animation-ready 3D geometry of a person as viewed in a single image.
Further, we compute a per-image lighting model to explain shading effects on 3D geometry and to enable relighting for realistic scene placement.

Our system utilizes the statistical body model GHUM~\cite{xu2020ghum}, which represents the human body $M(\cdot)$ as a parametric function of pose $\pose$ and shape $\shape$
\begin{equation}
    \small
    M (\shape, \pose): \pose \times{\shape} \mapsto \mat{V} \in \mathbb{R}^{3N} \label{eq:ghum}.
\end{equation}
GHUM returns a set of 3D body vertices $\mat{V}$. %
We also use imGHUM~\cite{alldieck2020imghum}, GHUM's implicit counterpart.
imGHUM computes the signed distance to the body surface $s^{body}_{\bx}$ for any given 3D point $\bx$ under a given pose and shape configuration: $\text{imGHUM}(\bx, \shape, \pose): \shape, \pose \mapsto s^{body}_{\bx}$.
Please refer to the original papers for details.

Given a monocular RGB image $\image$ (where the person is segmented), together with an approximate 3D geometry represented by GHUM/imGHUM parameters $\pose$ and $\shape$, our method generates an animation-ready, textured 3D reconstruction of that person. We represent the 3D geometry $\mathcal{S}$ as the zero-level-set of a signed distance field parameterized by the network $f$,
\begin{equation}\label{e:gM}
 \small
 \mathcal{S}_{\phi_f}=\left\{ \bx\in\real^3 \ \vert \ f(\bI, \pose, \shape, \bx;\phi_f)=(0, \ba)\right\},
\end{equation}
with learnable parameters $\phi_f$. In addition to the signed distance $s$ \wrt the 3D surface, $f$ predicts per-point albedo color $\ba$. 
In the sequel, we denote the signed distance and albedo for point $\bx$ returned by $f$ as $s_{\bx}$ and $\ba_{\bx}$, respectively. %
$\mathcal{S}$ can be extracted from $f$ by running Marching Cubes~\cite{marching_cubes} in a densely sampled 3D bounding box.
In the process of computing $s$ and $\ba$, $f$ extracts novel \textit{Structured 3D Features}.
In contrast to pixel-aligned features -- a popular representation used by other state-of-the-art methods -- our structured 3D features can be re-posed via Linear Blend Skinning (LBS), thus enabling reconstructions in novel poses as well as multi-image feature aggregation. We explain our novel Structured 3D Features in more detail below. %

\subsection{\Model~architecture}
\label{sec:arch}
We divide our end-to-end trainable network into two different parts. First, we introduce our novel Structured 3D Feature (S3F) representation. Second, we describe the procedure to obtain signed distance and color predictions.
Finally, we generalize our formulation to multi-view or multi-frame settings.

\vspace{1mm}
\noindent{\bf Structured 3D Features}. 
Recent methods have shown the suitability of local features for the 3D human reconstruction, due to their ability to capture high-frequency details.
Pixel-aligned features~\cite{saito2019pifu, saito2020pifuhd, huang2020arch, phorhum} are a popular choice, despite having certain disadvantages: (1) it is not straightforward to integrate pixel-aligned features from different time steps or body poses.
(2) a ray-depth ambiguity. In order to address these problems,
we propose a natural extension of pixel-alignment by lifting the features to the 3D domain: structured 3D features.
We obtain the initial 3D feature locations $\bv_i$ by densely sampling the surface of the approximate geometry represented by GHUM.
We initialize the corresponding features $\bff_i$ by projecting and pooling from a 2D feature map obtained from the input image $\bI$
\begin{equation}
    \small
    g: (\image, \pi(\bv_i)) \mapsto \bff_i\in\real^k \;, \label{eq:3d_feature_keys}
\end{equation}
where $\pi(\bv_i)$ is a perspective projection of $\bv_i$, and $g$ is a trainable feature extractor network.
The 3D features obtained this way approximate the underlying 3D body but not the actual geometry, especially for loose fitting clothing.
To this end, we propose to non-rigidly displace the initial 3D feature locations to better cover the human in the image.
Given the initial 3D feature locations $\bv_i$, we predict per-point displacements in camera coordinates using the network $d$
\begin{equation}
    \small
    \bv'_i = \bv_i + d(\bff_{i}, \be_{i}, \bv_i)  ,
    \label{eq:vertex_deformation}
\end{equation}
where $\be_i\in\real^{k}$ is a learnt semantic code for $\bv_{i}$, initialized randomly and jointly optimized during training.
Finally, we project the updated 3D feature locations again to the feature map obtained from $\bI$ (following the approach from Eq.~\ref{eq:3d_feature_keys}) obtaining our final structured 3D features $\bff'_i$. See Fig.~\ref{fig:diagram} for an overview.

\vspace{1mm}
\noindent{\bf Predicting geometry and texture}.
We continue describing the process of computing per-point albedo $\ba_{\bx}$ and signed distances $s_{\bx}$. 
We denote the set of structured 3D features as ($\bF'\in\real^{N \times k}, \bV'\in\real^{N\times3})$, where $\bF'$ refers to the feature vectors and $\bV'$ to their 3D locations, respectively.
First, we use a transformer encoder layer \cite{vaswani2017attention} to compute a master feature $\bff^{\star}_{\bx}$ for each query point $\bx$, using $\bV'$ as keys and $\bF'$ values, respectively.
From $\bff^{\star}_{\bx}$ we compute per-point albedo and signed distances using a standard MLP
\begin{equation}
    \small
    t: (\bx, \bV', \bF') \mapsto \bff^{\star}_{\bx} \mapsto s_{\bx}, \ba_{\bx}.
\end{equation}
While previous work maps query points $\bx$ to a continuous feature map (\eg pixel-aligned features), we instead aim to pool features from the discrete set $\bF'$.
Intuitively, the most informative features for  $\bx$ should be located in its 3D neighborhood.
We therefore first explore a simple baseline, by collecting the features of the three closest points of $\bx$ in $\bV'$ and interpolating them using barycentric coordinates.
However, we found this approach not sufficient (see Tab.~\ref{tab:ablation}).
With keys and queries based on 3D positions, we argue that a transformer encoder is a useful tool to integrate relevant structured 3D features based on learned distance-based attention.
In practice, we use two different transformer heads and MLPs to predict albedo and distance, please see Sup.\ Mat.\ for details.
Furthermore, we compute signed distances $s_{\bx}$ by updating the initial estimate of imGHUM with a residual,  $s_{\bx}=s^{body}_{\bx}+\Delta s_{\bx}$. This in practice makes the training more stable for challenging body poses.

To model scene illumination, we follow PHORHUM~\cite{phorhum} and utilize the bottleneck of the feature extractor network $g$ as scene illumination code $\bL$. We then predict per-point shading coefficient $\delta_{\bx}$ using a surface shading network $p$
\begin{equation}
    \small
     p(\bn_{\bx}, \bL; \phi_p) \mapsto \delta_{\bx},
\end{equation}
parameterized by weights $\phi_p$, where the normal of the input point $\bn_{\bx} = \nabla_{\bx} s_{\bx}$ is the estimated surface normal defined by the gradient of the estimated distance \wrt $\bx$.
The final shaded color is then obtained as $\bc_{\bx} = \delta_{\bx} \odot \ba_{\bx}$ with $\odot$ denoting element-wise multiplication.

\vspace{1mm}
\noindent{\bf 3D Feature Manipulation \& Aggregation}. 
Since the proposed approach relies on 3D feature locations originally sampled from GHUM's body surface, we can utilize GHUM to re-pose the representation.
This has two main benefits: (1) we can reconstruct in a pose different from the one in the image, \eg by resolving self-contact, which is typically not possible after creating a mesh, and (2) we can aggregate information from several observations as follows.
In a multi-view or multi-frame setting with $O$ observations of the same person under different views or poses, we define each input image as $\bI_t$, $\forall \ t \in \{ 1, \ldots, O \}$. 
Let us extend the previous notation to $(\bV'_t,\bF'_t)$ as structured 3D features for a given frame $t$.
To integrate features from all images, we use LBS to invert the posing transformation and map them to GHUM's canonical pose denoted as $\tilde\bV'$.
For each observation, we compute point visibilities $\bO_t\in\real^{N \times 1}$ (based on the GHUM mesh) and weight all feature vectors and canonical feature positions by the overall normalized visibility, thus obtaining an aggregated set of structured features
\vspace{-0.2cm}
\begin{align}
    \small
   \hat\bV' &= \sum_t^T \texttt{softmax}_t(O_t) \odot\tilde\bV'_t , \\
    \small  
   \hat\bF' &= \sum_t^T \texttt{softmax}_t(O_t) \odot\tilde\bF'_t
   \label{eq:aggregation}
\end{align}
where $\texttt{softmax}$ normalizes per-point contributions along all views. The aggregated structured 3D features $(\hat\bV',\hat\bF')$ can be posed to the original body pose of each input image, or to any new target pose. Finally, we run the remaining part of the model in the posed space to predict $s_{\bx}$ and $\ba_{\bx}$.

\subsection{Training \Model}
The semi-supervised training pipeline leverages a small subset of 3D synthetic scans together with images in-the-wild. During training, we run a forward pass for both inputs and integrate the gradients to run a single backward step.

\vspace{1mm}
\noindent{\bf Real data}. 
We use images in-the-wild paired with body pose and shape parameters obtained via fitting.
We supervise on the input images through color and occupancy losses.
Following recent work on neural rendering of combined signed distance and radiance fields \cite{volsdf}, we convert predicted signed distance values $s_{\bx}$ into density such that
\begin{equation}
    \small
    \sigma(\bx)=\beta^{-1}\Psi_\beta(-s_{\bx})\;,
\end{equation}
where $\Psi_\beta(\cdot)$ is the CDF of the Laplace distribution, and $\beta$ is a learnable scalar parameter equivalent to a sharpness factor.
After training, we no longer need $\beta$ for reconstruction but can use it to render novel views (see \cref{sec:novel_view}).

The color of a specific image pixel is estimated via the volume rendering integral by accumulating shaded colors and volume densities along its corresponding camera ray (see \cite{nerf} or Sup.\ Mat.\ for more details).
After rendering, we minimize the residual between the ground-truth pixel color $\bc_{\br}$ and the accumulation of shaded colors $\hat{\bc}_{\br}$ along the ray of pixel $\br$ such that $\mathcal{L}_{\text{rgb}} = \| \bc_{\br} - \hat{\bc}_{\br} \|_1$. Additionally, we also define a VGG-loss \cite{vgg_loss} $\mathcal{L}_{\text{vgg}}$ over randomly sampled front patches, enforcing the structure to be more realistic.
In addition to color, we also supervise geometry by minimizing the difference between the integrated density $\hat{\sigma}_{\br}$ and the ground truth pixel mask $\sigma_{\br}$: $\mathcal{L}_{\text{mask}} = \| \sigma_{\br} - \hat{\sigma}_{\br} \|_1$.

Finally, we regularize the geometry using the Eikonal loss~\cite{eikonal_loss} on a set of points $\Omega$ sampled near the body surface such that
\begin{equation}
    \small
    \mathcal{L}_{\text{eik}} = \sum_{\bx \in \Omega }(\|\nabla_{\bx} s_{\bx} \|_2 -1 )^2 .   
\end{equation}
The full loss for the real data is a linear combination of the previous components with weights $\lambda_{*}$ $\mathcal{L}_{\text{real}} = \mathcal{L}_{\text{rgb}} + \lambda_{\text{vgg}}\mathcal{L}_{\text{vgg}} + \lambda_{\text{mask}}\mathcal{L}_{\text{mask}} + \lambda_{\text{eik}}\mathcal{L}_{\text{eik}}$.

\begin{table}[t!]
\vspace{-0.3cm}
\centering
\resizebox{\columnwidth}{!}{%
\begin{tabular}{r c c c c c c c}
\cmidrule(lr){2-6}
& \multicolumn{3}{c}{Geometry} & \multicolumn{2}{c}{Color (PSNR)} \\
\cmidrule(lr){2-4} \cmidrule(lr){5-6}
Method & Chamfer $\downarrow$ & IoU $\uparrow$ & NC$\uparrow$  & Albedo $\uparrow$ & Shading $\uparrow$  \\
\cmidrule(lr){1-1} 
\cmidrule(lr){2-4} \cmidrule(lr){5-6} 
& \multicolumn{5}{c}{Architecture:} \\
\cmidrule(lr){1-1} 
\cmidrule(lr){2-4} \cmidrule(lr){5-6} 
\small{Using pixel-aligned features} & 3.59 & 0.360 & 0.828 & 13.17 & 14.13 \\
\small{+ Predicting SDF residual} & 5.43 & 0.428 & 0.851 & 13.19 & 13.06 \\
\small{+ Lifting features to 3D} & 0.606 & 0.644 & 0.765 & 9.82 & 10.07 \\
\small{+ Transformer} & 0.532 & 0.720 & 0.911 & \textbf{16.54} & 15.85 \\
\small{\textbf{FULL:} + Point displacement} & \textbf{0.339} & \textbf{0.734} & \textbf{0.924} & 16.31 & \textbf{16.67} \\
\cmidrule(lr){1-1} 
\cmidrule(lr){2-4} \cmidrule(lr){5-6}
& \multicolumn{5}{c}{Supervision regime:} \\
\cmidrule(lr){1-1} 
\cmidrule(lr){2-4} \cmidrule(lr){5-6}
\small{\textbf{FULL:} Only synthetic data} & 0.381 & 0.719 & 0.916 & 16.34 & 15.84 \\
\small{\textbf{FULL:} Only real data} & 0.444 & 0.723 & 0.907 & 10.96 & 13.59 \\
\cmidrule(lr){1-1} \cmidrule(lr){2-6}
\end{tabular}}
\caption{\small{{\bf Ablation of several of our design choices.}
Chamfer metrics are $\times10^{-3}$.}
}
\label{tab:ablation}
\vspace{-0.15cm}
\end{table}

\begin{table}[t!]
\vspace{-0.3cm}
\centering
\resizebox{\columnwidth}{!}{%
\begin{tabular}{r c c c c c c c}
\cmidrule(lr){2-6}
 & \multicolumn{3}{c}{Geometry} & \multicolumn{2}{c}{Color (PSNR)} \\
\cmidrule(lr){2-4} \cmidrule(lr){5-6}
Method & Chamfer $\downarrow$ & IoU $\uparrow$ & NC $\uparrow$ & Albedo $\uparrow$ & Shading  $\uparrow$ \\
\cmidrule(lr){1-1}
\cmidrule(lr){2-4} \cmidrule(lr){5-6}
\small{GHUM~\cite{xu2020ghum}} &  3.56 & 0.562 & 0.750 & - & -
\\ %
\small{PIFu~\cite{saito2019pifu}} & 6.61 & 0.519 & 0.738 & - & 11.15 \\
\small{Geo-PIFu~\cite{he2020geo}} & 9.61 & 0.453 & 0.710 & - & 10.62 \\
\small{PIFuHD~\cite{saito2020pifuhd}} & 4.94 & 0.552 & 0.749 & - & - \\
\small{PaMIR~\cite{zheng2021pamir}} & 5.35 & 0.597 & 0.763 & - & - \\
\small{ARCH~\cite{huang2020arch}} & 7.52 & 0.549 & 0.712 & - & 10.66 \\
\small{ARCH++~\cite{he2021arch++}} & 6.53 & 0.549 & 0.722 & - & 10.37 \\
\small{ICON~\cite{xiu2022icon}} & 3.53 & 0.622 & 0.785 & - & -  \\
\small{ICON~\cite{xiu2022icon} +~\cite{keypointnerf}} & 4.47 & 0.599 & 0.764 & - & - \\
\small{PHORHUM~\cite{phorhum}} & 2.92 & 0.594 & 0.814 & \underline{11.97} & 11.20 \\
\small{Ours (No shading)} & \underline{2.27} & \underline{0.675} & \underline{0.827} & - & \textbf{16.62} \\
\small{Ours} & \textbf{1.88} & \textbf{0.694} & \textbf{0.847} & \textbf{15.06} & \underline{14.81} \\
\cmidrule(lr){1-1} 
\cmidrule(lr){2-4} \cmidrule(lr){5-6}
\small{Ours (GT pose/shape)} & 0.339 & 0.734 & 0.924 & 16.31 & 16.67 \\
\cmidrule(lr){1-1} \cmidrule(lr){2-6}
\end{tabular}}
\caption{\small{{\bf Quantitative comparison against other monocular 3D human reconstruction methods}. Chamfer metrics are $\times10^{-3}$ and PSNR is obtained from all 3D scan vertices, including those not visible in the input image.}
\label{tab:monocular_reconstruction}}
\vspace{-0.02cm}
\end{table}

\begin{figure*}
\vspace{-0.3cm}
\includegraphics[width=\textwidth]{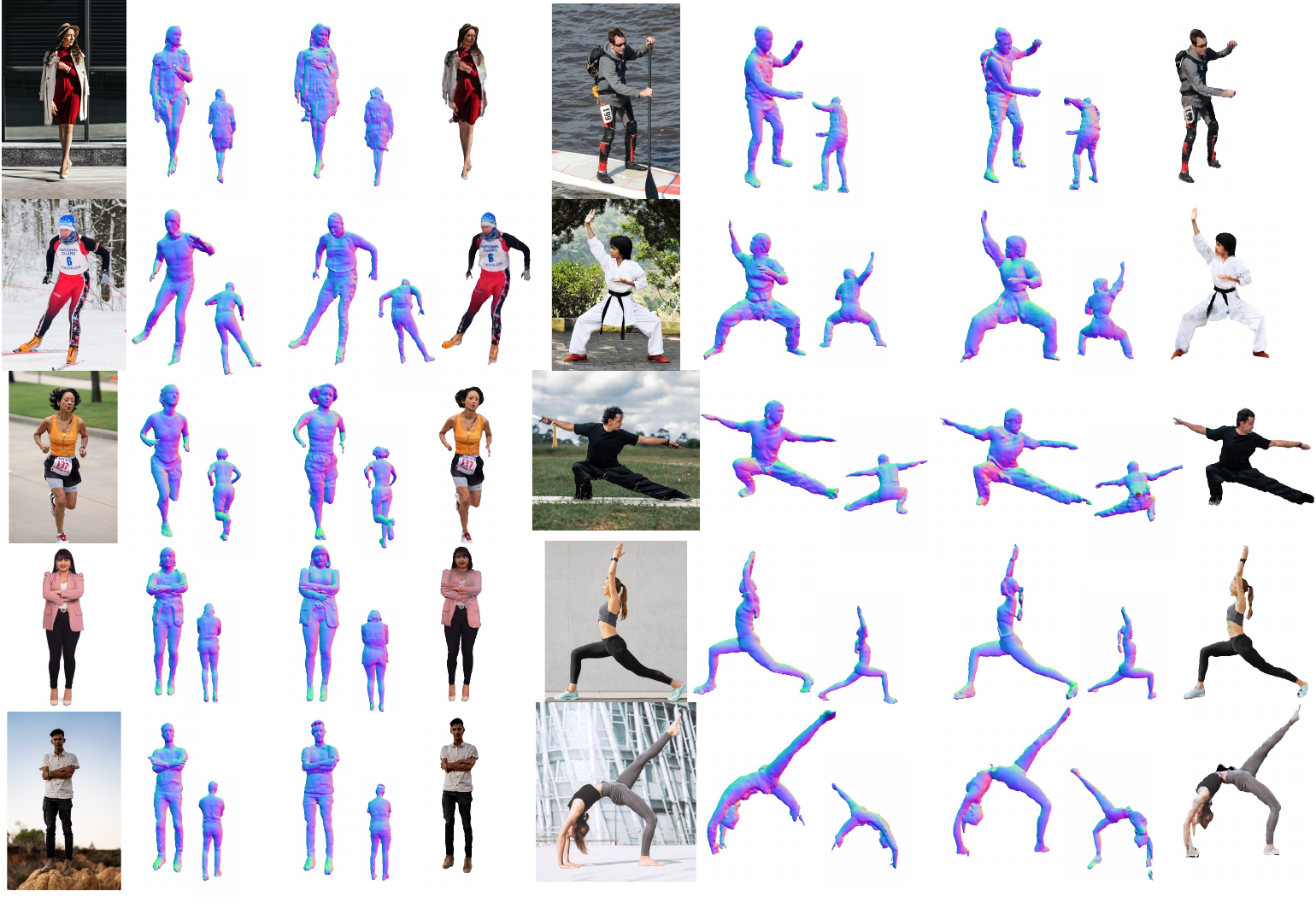}\\
\put(9,5){\scriptsize{Input image}}
\put(57,5){\scriptsize{ICON \cite{xiu2022icon}}}
\put(138,5){\scriptsize{OURS}}
\put(216,5){\scriptsize{Input image}}
\put(290,5){\scriptsize{ICON \cite{xiu2022icon}}}
\put(414,5){\scriptsize{OURS}}
\vspace{-0.12cm}
\caption{\small{\textbf{Qualitative 3D human reconstruction comparisons against the SOTA method ICON\cite{xiu2022icon}} for  complex body poses. Images on the right column feature challenging poses and viewpoints. Notice the additional detail provided by our method, especially for hair and faces. We additionally recover per-vertex albedo (not shown) and shaded color.
See Sup. Mat. for more examples and failure cases.}
\label{fig:reconstruction}}
\vspace{-0.18cm}
\end{figure*}

\vspace{1mm}
\noindent{\bf (Quasi-)Synthetic data}. Most previous work relies only on quasi-synthetic data in the form of re-rendered textured 3D human scans, for 3D human reconstruction.
While we aim to alleviate the need for synthetic data, it is useful for supervising features not observable in images in-the-wild, such as albedo, color, or geometry of unseen body parts.
To this end, we additionally supervise our model on a small set of pairs of 3D scans and synthetic images.
For the quasi-synthetic subset, we use the same losses as for the real data and add additional supervision from the 3D scans.
Given a 3D scan, we sample $B$ points on the surface together with its ground truth albedo $\ba_i^{\text{GT}}$ and shaded color $\bs_i^{\text{GT}}$.
We minimize the difference in predicted albedo and shaded color such that
$\mathcal{L}_{\text{3D rgb}} = \sum_{\bx \in B} \ \| \ba_{\bx}^{\text{GT}} - \hat{\ba}_{\bx} \|_1 + \| \bc^{\text{GT}}_{\bx} - \hat{\bc}_{\bx} \|_1$. 
We additionally sample a set of 3D points $\Omega$ close to the scan surface and compute inside/outside labels $l$ for our final loss, $\mathcal{L}_{\text{3D label}} = \sum_{\bx \in \Omega} \text{BCE}(l_{\bx}, \sigma(\bx))$,
where BCE is the Binary Cross Entropy function.
The overall loss for synthetic data $\mathcal{L}_{\text{synth}}$ is a linear combination of all losses described above.
We train our model by minimizing both real and synthetic objectives $\mathcal{L}_{\text{total}} = \mathcal{L}_{\text{real}} + \mathcal{L}_{\text{synth}}$.
Implementation details are available in the Sup.\ Mat.

\begin{figure}
\vspace{-0.32cm}
\includegraphics[width=\linewidth, trim={0.5cm 0 0.2cm 0}, clip=true]{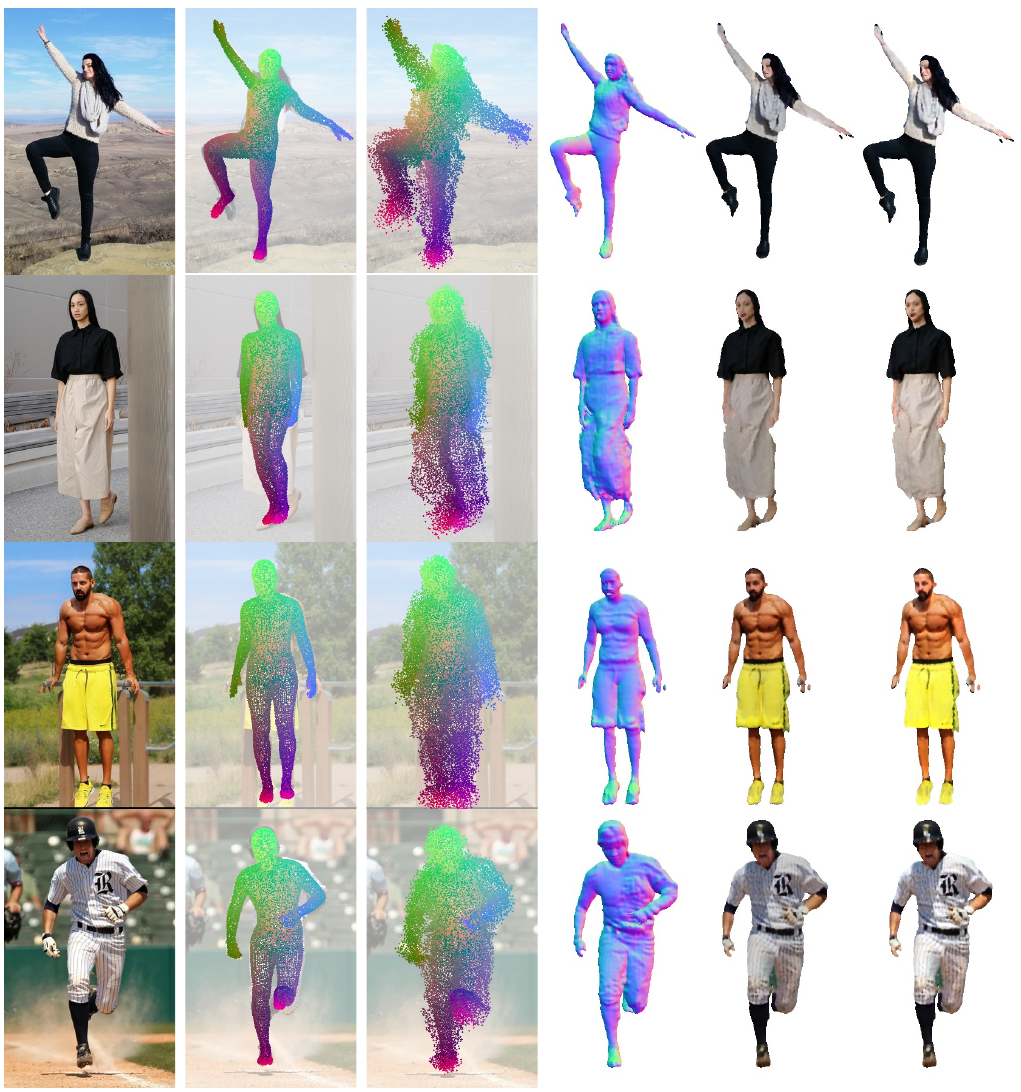}\\
\put(9,4){\scriptsize{Image}}
\put(40,4){\scriptsize{Initial points}}
\put(95,4){\scriptsize{S3Fs}}
\put(131,4){\scriptsize{Geometry}}
\put(175,4){\scriptsize{Albedo}}
\put(210,4){\scriptsize{Shaded}}
\vspace{-0.1cm}
\caption{\small{\textbf{Output of \Model.} Given an input image (first col.) and GHUM pose estimates (second col.), \Model~displaces points efficiently to cover relevant features from the image (third col.). Displaced points cover areas that  not necessarily belong to the foreground, but tend to represent loose clothing (\eg hair and loose clothing in first and second examples respectively). This coverage and the efficient feature retrieval allows to correct potential errors in the body such as the feet in the first two examples or the overall body orientation in the third row. Zoom-in recommended.}
\label{fig:ours}}
\vspace{-0.05cm}
\end{figure}

\begin{figure}
\vspace{-0.1cm}
\includegraphics[width=1\linewidth, trim={0 0.0cm 0 0}, clip=true]{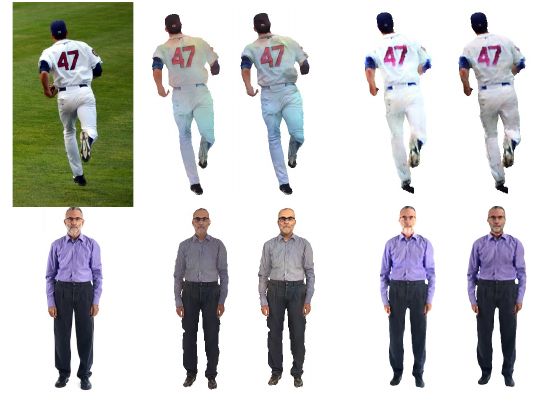}\\
\put(13,9){\scriptsize{Input Image}}
\put(73,9){\scriptsize{Albedo}}
\put(112,9){\scriptsize{Shaded}}
\put(163,9){\scriptsize{Albedo}}
\put(203,9){\scriptsize{Shaded}}
\put(80,0){\scriptsize{PHORHUM \cite{phorhum}}}
\put(183,0){\scriptsize{OURS}}
\vspace{-0.1cm}
\caption{\small{{\bf Qualitative comparison against the state-of-the-art method PHORHUM\cite{phorhum}} for albedo and shading estimation from monocular images.}
\label{fig:shading_estimation}}
\vspace{-0.22cm}
\end{figure}

\vspace{-0.15cm}
\section{Experiments}\label{sec:exp}

Our goal is to design a single 3D representation and an end-to-end semi-supervised training methodology that is flexible enough for a number of tasks.
In this section, we evaluate our model's performance on the tasks of monocular 3D human reconstruction, albedo, and shading estimation, as well as for novel view rendering and reconstruction from both single-images and video. Finally, we illustrate our model's capabilities for 3D garment editing.

\subsection{Data}
We follow a mixed supervision approach by combining synthetic data with limited clothing or pose variability, with in-the-wild images of people in challenging poses/scenes.

\vspace{1mm}
\noindent{\bf Synthetic data}. We use 35 rigged and 45 posed scans from RenderPeople~\cite{renderpeople} for training. We re-pose the rigged scans to 200 different poses and render them five times with different HDRI backgrounds and lighting using Blender~\cite{blender}. With probability $0.4$ we use a frontal view, otherwise we render the person from a random azimuth and a uniform random elevation in $\left[ -20, 20 \right]^{\circ}$. We obtain ground-truth albedo directly from the scan's texture map, and shade the full scan (including occluded regions) to obtain ground-truth shaded colors before rendering.

\vspace{1mm}
\noindent{\bf Images in-the-wild}. We use the HITI dataset~\cite{bazavan2021hspace}, a collection of images in-the-wild with 2D key-point annotations and human segmentation masks, to which we fit GHUM by minimizing the 2D joint reprojection error. %
We automatically remove images with high fitting error,
resulting in 40K images for training.
We use the provided test set to validate results.
More details are provided in the Sup. Mat.

\vspace{1mm}
\noindent{\bf Evaluation data}. We compare our method for monocular reconstruction using a synthetically generated test set based on a test-split of 3D scans, featuring significant diversity of body poses (\eg running), illumination, and viewpoints.
For images in-the-wild, we obtain GHUM parameter initialisation using~\cite{grishchenko2022blazepose} and further optimize by minimizing 2D joint reprojection errors.
For the task of novel-view rendering we use GHS3D~\cite{xu2020ghum}, which consists of 14 multi-view videos of different people, between 30-60 frames each, with 4 views for training and 2 for testing.
We evaluate on test views and compare results based on either a single training image or full video. Note that our method does not "train", but only computes a forward pass based on a training image. %

\subsection{Monocular 3D Human Reconstruction}

\noindent{\bf Ablation Study}. We ablate our main methodological choices in Table~\ref{tab:ablation}. We use ground-truth pose and shape parameters for the test set, to factor out potential errors in body estimation that otherwise dominate the reconstruction metrics.
We provide different baselines to demonstrate the contribution at each step. Albedo and shading evaluation is based on nearest neighbors of the scan, including the back or non-visible regions, which tend to dominate PSNR error rates.
In the first two rows we show results of pixel-aligned only baselines.
By lifting features to the 3D domain and placing them around the body, our method already achieves competitive performance.
This is in line with other works \cite{xiu2022icon, huang2020arch, he2021arch++} which showed that extending a statistical body model leads to better reconstruction for challenging poses and viewpoints.
We improve on this by using a transformer architecture associating query points with 3D features, %
especially when using our S3F (full model). 
Finally, we compare our performance when training only on real or synthetic data and show that we get the best results when using a mix of both. Moreover, real data helps generalization to in-the-wild images, not captured in our test set.

\noindent{\bf SOTA Comparison}. 
We evaluate our method for monocular 3D human reconstruction and compare with previous methods quantitatively in Table~\ref{tab:monocular_reconstruction}.
Here, we do not assume ground-truth pose/shape for each image, and instead obtain them from the input images alone, for all methods.
We observe that while metrics are dominated by errors in pose estimation, our method is more robust and can correct these errors by using S3Fs.
Additionally, our model recovers albedo and shading significantly better than baselines.

Qualitatively, we compare against ICON~\cite{xiu2022icon} in Fig.~\ref{fig:reconstruction}, for different levels of body pose complexity. Our method is more robust to both challenging poses and loose clothing than previous SOTA while additionally predicting color, being animatable and relightable. We provide additional comparisons to other methods in Sup. Mat.
We also compare our method qualitatively against PHORHUM~\cite{phorhum} on the task of albedo and shading estimation from monocular images in Fig.~\ref{fig:shading_estimation}. We observe that our method is more robust and we hypothesize this is due to PHORHUM's trained on synthetic data only, 
which impacts its ability to generalize to images in-the-wild.
We show more results of our method in Fig.~\ref{fig:ours}, where
we also visualize the initial and displaced 3D locations of our body points. While the initial GHUM fit might lead to noisy estimates, the displacement step is able to correct errors (feet in first \& second rows) and better cover loose clothing (second row).

\begin{table}[t!]
\vspace{-0.3cm}
\centering
\setlength{\tabcolsep}{2pt}
\resizebox{1.0\linewidth}{!}{%
\begin{tabular}{r c c c c c c c}
\cmidrule(lr){2-8}
& Chamfer$\downarrow$ & IoU$\uparrow$ & NC$\uparrow$ & PSNR$\uparrow$ & SSIM$\uparrow$ & LPIPS$\downarrow$ & Train time \\
\cmidrule(lr){1-1} \cmidrule(lr){2-8}
\small{NeuralBody \cite{neuralbody}} & 0.790 & 0.887 & 0.810 & 24.70 & 0.829 & 0.236 & hours \\
\small{H-NeRF \cite{xu2021h}} &  0.218 & 0.932 & 0.890 & 24.92 & 0.852 & 0.232 & hours \\
\small{Ours (finetuned)} & 0.244 & 0.841 & 0.910 & 27.54 & 0.871 & 0.148 & 30 min\\
\cmidrule(lr){2-8}
& \multicolumn{7}{c}{One-shot novel view rendering} \\
\cmidrule(lr){2-8}
\small{Ours (1 image)} & 0.504 & 0.795 & 0.899 & 23.75 & 0.825 & 0.167 & 0 \\

\small{Ours (video)} & 0.473 & 0.807 & 0.905 & 24.82 & 0.836 & 0.148 & 0 \\
\cmidrule(lr){1-1} \cmidrule(lr){2-8}
\end{tabular}}
\caption{\small{{\bf Quantitative comparison against recent novel view rendering methods for the GHS3D Dataset\cite{xu2021h}}. We achieve competitive reconstruction, and view rendering metrics using just one forward pass, especially when taking multiple frames as input. This validates our proposed feature integration scheme.}
\label{tab:novel_view_rendering}}
\vspace{-0.05cm}
\end{table}
\begin{figure}
\centering
\vspace{-0.3cm}
\includegraphics[width=0.94\linewidth, trim={0 2.8cm 0 0}, clip=true]{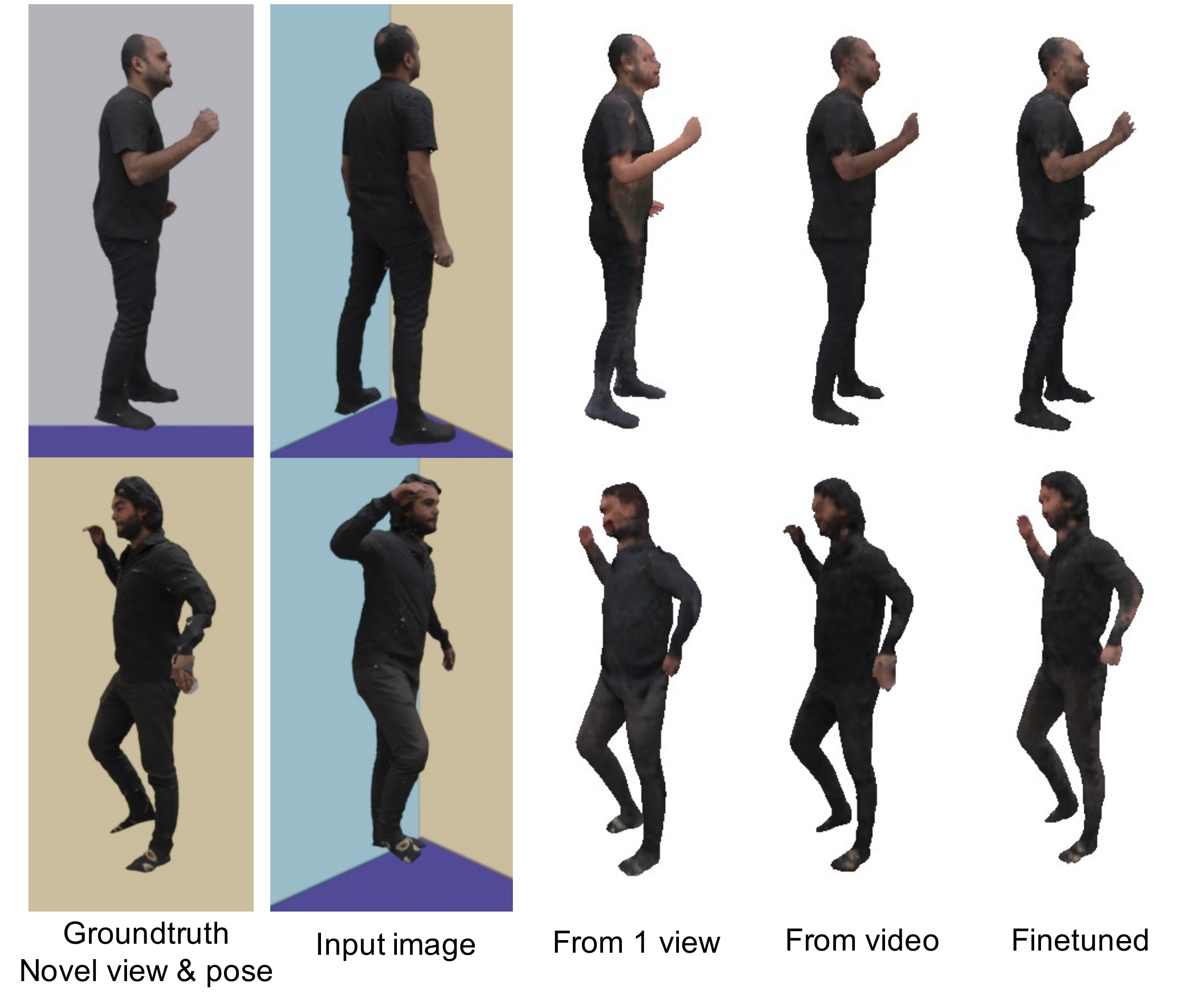}
\put(-217,-5){\scriptsize{GT view \& pose}}
\put(-165,-5){\scriptsize{Input Image}}
\put(-122,-5){\scriptsize{From 1 view}}
\put(-80,-5){\scriptsize{From video}}
\put(-39,-5){\scriptsize{Finetuned}}
\caption{\small{\textbf{3D Human reconstruction from one image, video and after finetuning} our network geometry and color heads for 30 minutes. 
The image used as input in the single-view case is shown in the second column. By using only this information, previously occluded body areas might lead to larger errors when rendering from novel viewpoints or in a different pose, \eg first example, where face and frontal details need to be hallucinated. This can be corrected by integrating information from multiple frames.}
\label{fig:one_frame_vs_multiple}
}
\vspace{-0.05cm}
\end{figure}
\subsection{Applications}
\label{sec:novel_view}

The main applications of human digitization (\eg VR/AR or telepresence) require considerable control over virtual avatars, in order to animate, relight and edit them. 
Here, we evaluate the performance of our approach for novel view rendering to a full video, given as little as one input image. This process entails the ability to re-pose reconstructions and integrate features from different views or video frames. We then showcase the model's potential for relighting and clothing editing.

\begin{figure}
\vspace{-0.15cm}
\includegraphics[width=\linewidth]{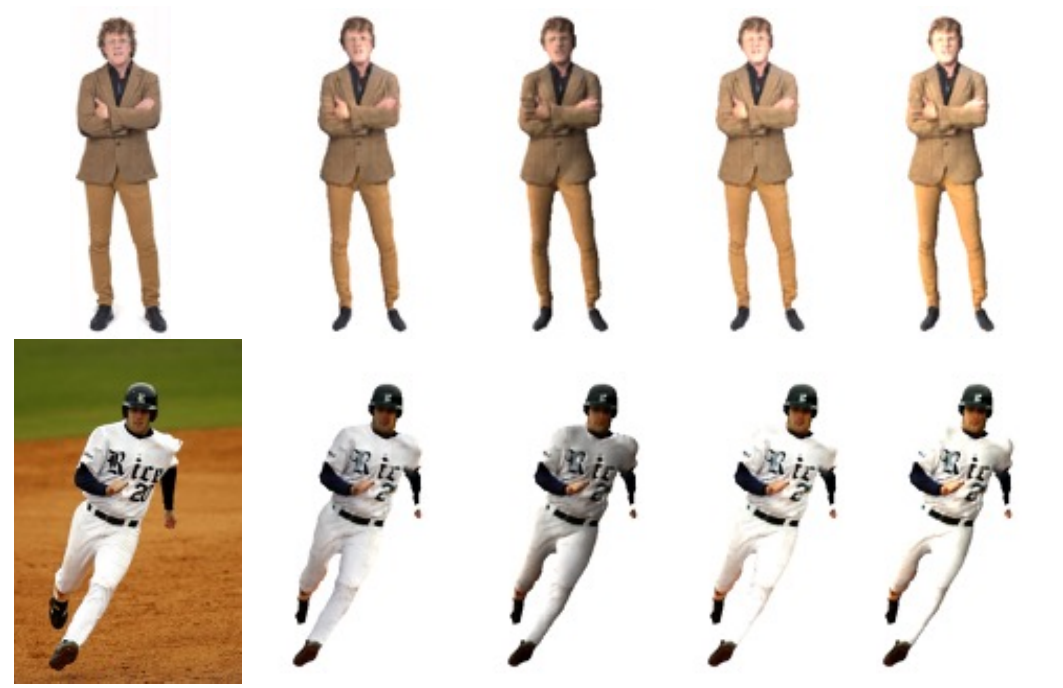}\\
\put(7,4){\small{Input image}}
\put(100,4){\small{Relighted Reconstructions}}
\vspace{-0.08cm}
\caption{\small{\textbf{Qualitative results on 3D Human Relighting}.
}
\label{fig:shading}}
\vspace{-0.15cm}
\end{figure}

\noindent{\bf Novel view synthesis}. 
For novel-view synthesis, previous work often requires tens of GPU hours to train person-specific networks. Here, we compare against two such recent works, NeuralBody \cite{neuralbody} and H-Nerf \cite{xu2021h}, on the GHS3D dataset~\cite{xu2021h}.
We report reconstruction and novel view metrics in Table~\ref{tab:novel_view_rendering}. These show that our method obtains comparable performance without retraining, just from a forward pass.
For video, we have enough supervision to robustly fine-tune the geometric head of our network, in just 30 minutes, to obtain a significant performance boost in both geometry and rendering metrics. In Fig.~\ref{fig:one_frame_vs_multiple} we show the difference between reconstructing from a single view or full video. Even though our method yields a reasonable 3D reconstruction from just a single image, it has to hallucinate color in occluded areas (\eg face and front body parts in the first row). The resulting inconsistency can be largely resolved when novel views are considered. 

\begin{figure}
\vspace{-0.25cm}
\includegraphics[width=\linewidth]{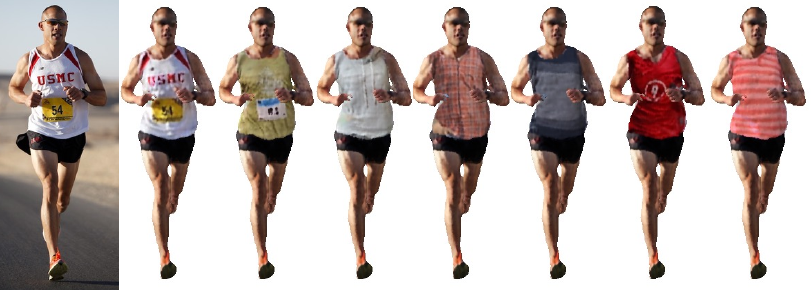}\\
\put(7,8){\scriptsize{Image}}
\put(30,8){\scriptsize{Reconstruction}}
\put(110,8){\scriptsize{Cloth texture transfer}}
\vspace{-0.1cm}
\caption{\small{\textbf{Clothing texture transfer}.
Given an image of a target person, we identify S3Fs that project inside upper-body cloth segmentation~\cite{li2020self}, and replace their feature vectors for those obtained from other subjects.
More examples, cloth types and reference images and segmentation masks are shown in Sup.\ Mat.
}
\label{fig:editing}}
\vspace{-0.25cm}
\end{figure}
\noindent{\bf Relighting}. By predicting per-point albedo and a scene illumination, our model returns a relightable reconstruction of the person in the target image. We show a number of relighted reconstructions in Fig.~\ref{fig:shading} given a target input image (left column).
The reconstructions are consistently re-shaded in novel scenes, thus enabling 3D human compositing applications.

\noindent{\bf Clothing editing}. We harvest the semantic properties of the proposed Structured 3D Features to explore potential applications for 3D virtual try-on. Given an input image of a person and a segmentation mask~\cite{li2020self} of a particular piece of clothing (\eg upper-body), we first find body points that project inside the target mask. By exchanging their features with those obtained from other images, we can effectively transfer clothing texture.
Fig.~\ref{fig:editing} showcases diverse clothing editing cases for the upper-body. See Sup.\ Mat.\ for other subjects and garments. 
While 3D clothing transfer is a complex problem by itself~\cite{mir20pix2surf,xu20213d}, we tackle a much more challenging scenario than previous methods that typically map color to a known 3D template; and we use a single, versatile end-to-end model for monocular 3D reconstruction, relighting, and 3D editing.

\vspace{-0.15cm}
\section{Discussion}
\noindent{\bf Limitations}. 
Failures in monocular human reconstruction occur mainly due to incorrect GHUM fits or very loose clothing. 
The lack of ambient occlusions in the proposed scheme might lead to incorrect albedo estimates for images in-the-wild. Misalignment in estimated GHUM pose \& shape might lead to blurriness, especially for faces. We show failure cases in Supp. Mat.

\vspace{1mm}
\noindent{\bf Ethical considerations}. 
We present a human digitization tool that allows relighting and re-posing avatars. However, neither is it intended, nor particularly useful for any form of deep fakes, since the quality of reconstructions is not at par with facial deep fakes.
We aim to improve model coverage for diverse subject distributions by means of a weakly supervised approach and by taking advantage of extensive image collections, for which labeled 3D data may be difficult or impossible to collect. %

\vspace{1mm}
\noindent{\bf Conclusion}.
We have presented a controllable transformer methodology, \Model, relying on versatile attention-based features, that enables the 3D reconstruction of rigged and relightable avatars by means of a single semi-supervised end-to-end trainable model. 
Our experimental results illustrate how S3F models obtain state-of-the-art performance for the task of 3D human reconstruction and for albedo and shading estimation. We also show the potential of our models for 3D virtual try-on applications.

\section*{Supplementary Material}
\appendix

In this supplementary, we describe the implementation details of our method extensively and provide more results and failure cases. We also include a Supplementary Video summarizing our contributions and results.

\section{Implementation Details}\label{sec:implem}

\vspace{1mm} \noindent{\bf Data.} 
Our synthetic data is based on a set of 3D RenderPeople Scans\cite{renderpeople}. We use 35 rigged and 45 posed scans for training. The rigged scans are re-posed to 200 different poses sampled randomly from the CMU motion sequences~\cite{cmu_mocap}. With probability $0.4$ we render a scan from a frontal view, otherwise, we render from a random azimuth and a uniform random elevation in $\left[ -20, 20 \right]^{\circ}$.
We render each scan using high dynamic range image (HDRI)~\cite{hdri} lighting and backgrounds using Blender~\cite{blender}.
We obtaining ground-truth albedo directly from the scan's texture map and bake the full scan's shading (including occluded regions) to obtain ground-truth shaded colors.

For real images, we use the HITI dataset~\cite{bazavan2021hspace}, which contains 150K images in-the-wild with predicted foreground segmentation masks and annotated 2D human keypoints. We obtained initial GHUM parameters for each image by estimating pose and shape using~\cite{grishchenko2022blazepose}. We then further optimized pose and shape parameters by minimizing the 2D reprojection error, the normalizing flow pose prior from~\cite{xu2020ghum,zanfir2020weakly} and a body shape prior. The weights for joint reprojection, body pose regularization and body shape regularization are $10$, $1$ and $10$ respectively, and we assumed a perspective camera projection with fixed focal length.
After fitting, we remove fits that have an average reprojection error greater than 3 pixels or where at least 10\% of the body surface projects outside the segmentation mask, leading to 40k training images. For inference on images-in-the-wild we follow the same fitting procedure, by leveraging predicted 2D keypoints instead of ground-truth annotations.

\vspace{1mm} \noindent{\bf Architecture.} 
We take masked input images at $512\times512$~px resolution. 
We augment the input with rendered normal and semantic maps to provide information about the GHUM fit to the feature extractor network.
The semantic map is obtained by rendering the original template vertex locations as vertex colors, essentially defining a dense correspondence map to GHUM's zero-pose.
We normalize both maps between 0 and 1 and stack them with the original image before passing all to the image feature extractor network.
We noticed that by concatenating normal and semantic maps, the network is better able to correct noisy GHUM fits and their geometry.
The feature extractor network is a U-Net~\cite{unet} with 6 encoder and 7 decoder layers with sizes $\left[64,128,256,512,512,512\right]$ and $\left[512,512,512,512,256,256,256\right]$ respectively. 
The illumination code is extracted from the bottleneck and has shape $8 \times 8 \times 512$.
The output per-pixel feature maps is of shape $512 \times 512 \times 256$, with $256$-feature vectors.

We next detail how we sample points on the body surface and pool from image features.
We explored different point densities sampled from the body surface, all based on the original GHUM mesh to maintain correspondences.
To subdivide a mesh, we add a vertex in the center of each edge, increasing the number of faces by a factor of 4. 
The subdivision is fast to perform at train/test time and does not cause any significant overhead.
However, naively subdividing points on the mesh leads to a large number of body points causing memory issues in later stages (\eg processing them on the transformer encoder), thus we additionally run K-Means on the  subdivided template mesh obtaining clusters of 2k, 5k, 8k, 10k, 12k, 15k and 18k body points.
We ablate qualitatively the number of points on the model in Fig.~\ref{fig:num_points_supp}.
Using fewer points does not significantly affect the results, but produces blurrier color reconstructions.
Our final model uses 18k points.
During inference, given the GHUM fit of an image and estimated camera parameters, we project these points to the image and extract image features. We use the last 64 features of the previously extracted image features to predict per-vertex deformation, using a 2-Layer MLP with hidden shapes $\left[64, 3\right]$ and leaky-ReLU after the first layer. We project the deformed vertices again to obtain the remaining 192 per-pixel features.

The goal of the transformer encoder
is to efficiently map a query point $\bx$ to the Structured 3D Features. We first map both deformed body points and query point positions to a higher-dimensional space using positional encoding with 6 frequencies, and apply a shared 2-Layer MLP with output size 256.
In practice, we use two MLPs predicting geometry and albedo independently.
We next apply attention~\cite{vaswani2017attention} to combine per-point features and obtain $\bff^{\star}_{\bx}$. 
The final geometry and color heads are both MLPs with eight 512-dimensional fully-connected layers and Swish activation~\cite{swish}, an output layer with Sigmoid activation for the color component, and a skip connection to the fourth layer. The shading network $s$ is conditioned on the previously extracted illumination code and consists of three 256-dimensional fully-connected layers with ReLU activation, including the output layer.

The weights of all layers are initialized with Xavier initialization~\cite{glorot}, with the $\beta$ parameter (Eq. 9) being initialized as 0.1. 
We train all network components jointly end-to-end for 500k iterations using the Adam optimizer~\cite{adam}, with an initial learning-rate of $1 \times 10^{-4}$ that linearly decays with factor 0.9 every 50k steps. Training takes 5 days under 8 GPUs V100 at batch size 8. 
During training, the time bottleneck is the U-Net feature extractor. However, the transformer is the memory bottleneck when computing attention, which scales with the number of body points and query points.
The 3D reconstructions are obtained after running Marching Cubes~\cite{marching_cubes} at a densely sampled volume of $512 \times 512 \times 512$ points, as common in previous works~\cite{saito2019pifu,phorhum,saito2020pifuhd}. After running Marching Cubes for geometry reconstruction, we query the network again on the reconstruction's vertices, to obtain albedo and color estimations and texture the mesh.
To render new views, we explored the possibility of rendering the colored mesh vs. using neural rendering without obtaining the 3D mesh. We did not find significant differences between the two. 
In our results, we apply the former and first obtain the mesh with Marching Cubes for convenience. %

\vspace{1mm} \noindent{\bf Point and pixel sampling strategy.}
For training on the 3D RenderPeople scans, we sample 128 points uniformly on the scan's surface, and 128 points close to the body (by adding noise $\sim \mathcal{N}(0, 1~\text{cm})$ to scan vertices). %
For the Eikonal Loss, we sample points around GHUM by adding noise $\sim \mathcal{N}(0, 10~\text{cm})$ to GHUM's original vertices.

We rely on the original segmentation mask to compute image-based losses, starting by sampling pixels efficiently on foreground regions to let the network focus on occupied regions. We randomly sample 32 pixels from the input image, from which 75\% are sampled inside the foreground mask, and the rest are outside the mask. 
For the VGG-loss~\cite{vgg_loss} $\mathcal{L}_{\text{vgg}}$ we render a $16 \times 16$ patch by sampling its center pixel randomly from the foreground mask.

\vspace{1mm} \noindent{\bf Loss functions.}
We next describe how we obtain pixel colors and our neural rendering losses more in detail. 
Following the notation of the main document, the color of the pixel $\mat{c}_k$ is approximated by a discrete integration between near and far bounds $t_n$ and $t_f$ of a camera ray $\mat{r}(t) = \mat{o} + t\mat{d}$ with origin $\mat{o}$:
\begin{equation}
    \bc_k = \int_{t_n}^{t_f} T(t) \sigma(r(t))\mat{c}(\mat{r}(t), ) dt\,,
\end{equation}
where:
\begin{equation}
    T(t) = \exp \left( -\int_{t_n}^{t}\sigma(\mathbf{r}(s))ds \right)\,,
\end{equation}

and $\sigma(\cdot)$ is the predicted density of a query point, as defined in the main document.
Note that the point color being integrated is the shaded point color, which is a composition of its albedo and normal, obtained from $\bn_{\bx} = \nabla_{\bx} s_{\bx}$ for point $\bx$. In practice, for each target pixel, we sample 64 points uniformly along the ray, within the bounding box of GHUM's nearest and furthest $z$ vertex locations padded with 10~cm.

Our experience is that the main trade-off during training is on excessive detail (noise in geometry) vs. excessive smoothness and found the weight of the Eikonal component to be the most important hyperparameter to balance the capacity of the model to generate wrinkles. In this manner, we run a hyperparameter sweep on $\lambda_{\text{eik}}$ from 0.01 to 0.5. The loss weights for the final model are $\lambda_{\text{rgb}}=10$, $\lambda_{\text{vgg}}=30$, $\lambda_{\text{mask}}=5$, $\lambda_{\text{eik}}=0.1$, 
$\lambda_{\text{3D rgb}}=200$, $\lambda_{\text{3D label}}=30$.
Losses $\mathcal{L}_{\text{synth}}$ and  $\mathcal{L}_{\text{real}}$ contribute equally to the total loss.

\vspace{1mm} \noindent{\bf Negative results.} 
We explored a range of ideas that ended up degrading or not affecting performance in our setting, and we briefly report them here. Our intention in reporting these results is to save time for future research and to give a more complete picture of our attempts. Note that these results are specific to our particular setup and are not meant to discourage potentially fruitful avenues of research.
\begin{enumerate}
    \item Discriminator. We explored adding an additional adversarial loss~\cite{goodfellow2020generative,mescheder2018training} to obtain more realistic details, specially from non-visible viewpoints. Nevertheless, the discriminator was leading to noise in geometry and incorrect wrinkles, and was often unstable during training. In practice, we obtained better results in geometry by just minimizing $\mathcal{L}_{\text{3D label}}$ on the real 3D scans.
    \item Hessian Loss~\cite{implicit_hessian_loss}. We explored the possibility of regularizing the predicted Signed Distance Field by encouraging small second-order derivatives with respect to input points, but this was highly memory consuming and increased training time. We also did not observe significant changes with respect to regularizing first-order derivatives via an Eikonal loss~\cite{eikonal_loss}.
    \item Perceiver~\cite{jaegle2021perceiver}. Since the transformer architecture is the memory bottleneck, we explored reducing the dimensionality of the Structured 3D Features before assigning features to query points. We found, however, that this was leading to very smooth geometry and texture. The intuition after this is that denser local features are better suited for 3D reconstruction of high-frequency details.
    \item Canonicalization. Our initial attempts explored canonicalization as a way to simplify the learning problem and efficiently utilize the body prior by learning a single model in T-Pose.
    However, we observed that the canonicalization step is very sensitive to small GHUM errors, resulting in a lack of high frequencies like wrinkles in geometry or details in texture. Therefore we use our transformer in the posed space and map to canonical space only when re-posing, by using skinning weights after collecting per-point features.
    \item 3D normal supervision on synthetic data.
    We explored having an additional loss function to minimize the difference between the predicted point normals and groundtruth 3D scan normals, evaluated on on-surface points. However, this led to smoother geometry, performing similarly to an additional Eikonal regularization. 
    We tried unit normalizing $\nabla_{\bx} s_{\bx}$ before supervision to have this loss function specifically focused on the normal direction instead of the magnitude, though obtaining similar results.
    \item Additional loss function to penalize the difference between imGHUM's SDF and our predicted SDF. This acts as a regularization for the learnt SDFs, but was tying the network excessively to the body and reduced its capacity to generate loose clothing or recover from noisy GHUM fits.
    \item Pose-dependent SDF estimations. We briefly explored the possibility of having an additional MLP that takes GHUM pose/shape and predicts per-point deformation (both deforming query points or deforming feature points). The intuition after this is to try to learn additional pose-dependent effects within training, by also training on a significant pose diversity. However, the MLP converged to predict negligible displacements and we suspect that tackling this task effectively requires tailored data and objective functions.
\end{enumerate}

\begin{figure}[t]
\centering
  \includegraphics[width=\linewidth]{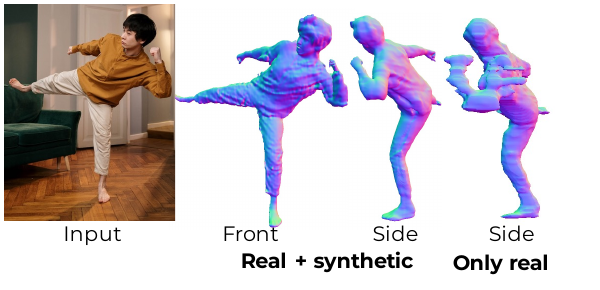}
  \caption{\label{fig:rebuttal_only_real_full_supp}
  Results of the full method (trained on both real and synthetic data) in comparison to a method trained only from real images. As observed in this example, training with no synthetic supervision leads to significant uncertainty in the z-axis during reconstruction. The body template is useful to regularize the process but the method becomes unable to generate realistic 3D reconstructions.
  }
\end{figure}

\begin{figure}[t]
\centering
  \includegraphics[width=\linewidth]{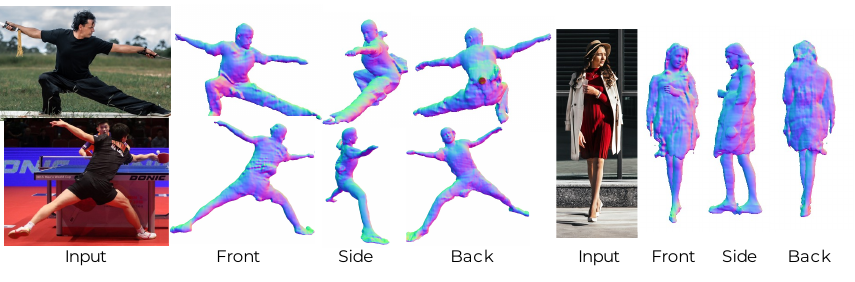}
  \caption{\label{fig:rebuttal_sideview_supp}
  Examples showing the front, side and back views of the geometry reconstructed from our method. The proposed approach tackles extreme poses (left) or people with loose clothing (right).  
  }
\end{figure}

\section{Additional Results}\label{sec:exp_supp}

In this section, we provide more analysis and results with respect to the experiments in the main paper. Figure~\ref{fig:rebuttal_only_real_full_supp} shows a reconstruction of the full method trained on both real and synthetic data, in comparison to a method trained only on real images. We here show side views of the reconstructions to show the consistency of geometry in all views. In contrast, when training only on real images, the method is unable to properly solve the uncertainty along the Z-axis and often generates artifacts. Figure~\ref{fig:rebuttal_sideview_supp} depicts more results from the full method, including front, side and back views for different images. On the left, we show examples of extreme poses and the example on the right depicts a person with loose clothing. Note that monocular 3D reconstruction methods are inherently ambiguous \wrt depth, and the use of synthetic data and pose priors during pose estimation limits some of the ambiguities. Our multi-view extension further alleviates the problem.

\begin{figure*}
\includegraphics[width=\linewidth]{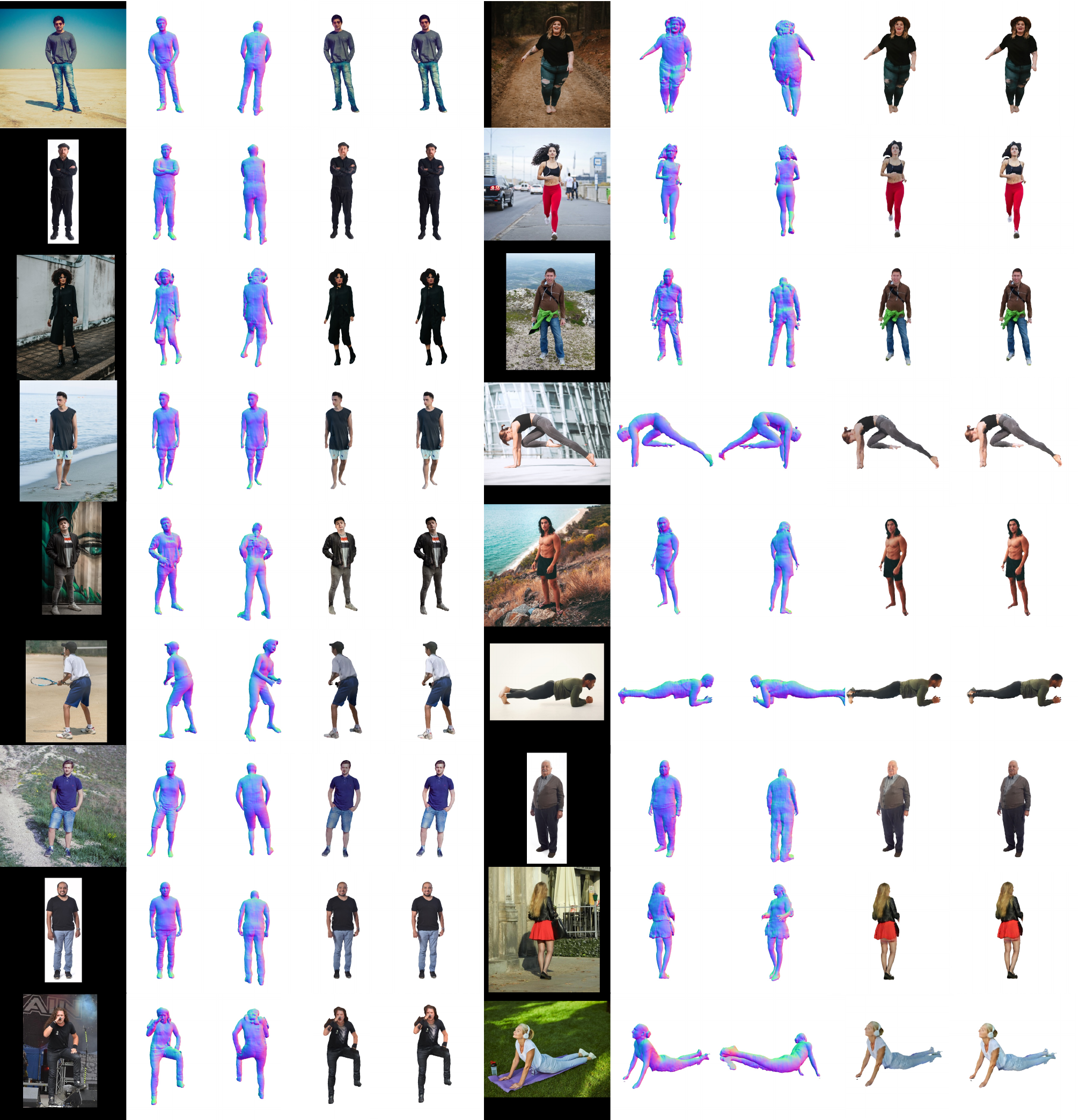}\\
\put(8,4){\small{Input image}}
\put(69,4){\small{Front}}
\put(106,4){\small{Back}}
\put(141,4){\small{Albedo}}
\put(182,4){\small{Shading}}
\put(231,4){\small{Input image}}
\put(293,4){\small{Front}}
\put(348,4){\small{Back}}
\put(398,4){\small{Albedo}}
\put(454,4){\small{Shading}}
\caption{\small{\textbf{Additional qualitative examples from our method}.
}
\label{fig:qualitative_results1_supp}}
\end{figure*}

\begin{figure*}
\includegraphics[width=\linewidth]{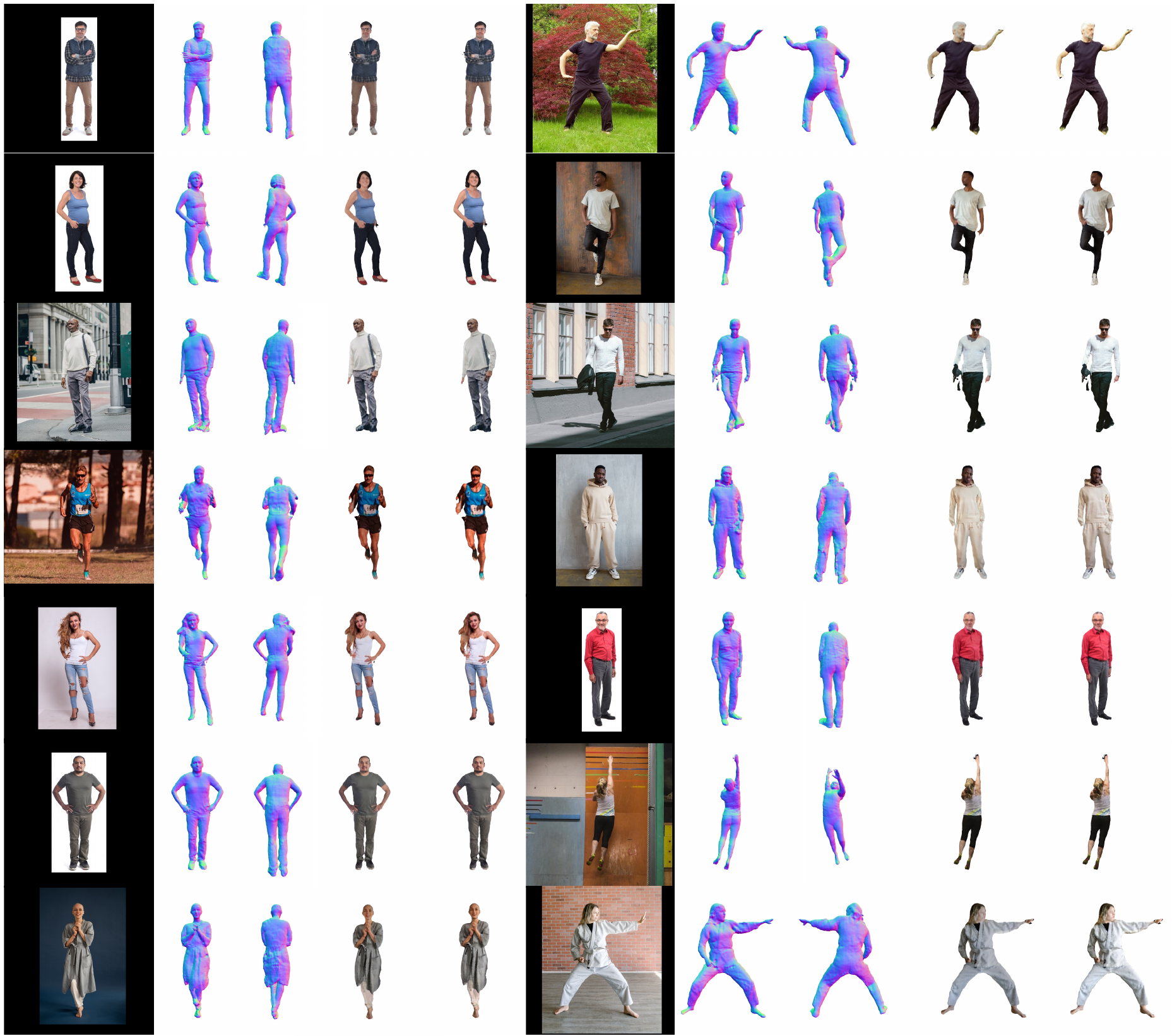}\\
\put(11,4){\small{Input image}}
\put(74,4){\small{Front}}
\put(109,4){\small{Back}}
\put(139,4){\small{Albedo}}
\put(185,4){\small{Shading}}
\put(230,4){\small{Input image}}
\put(298,4){\small{Front}}
\put(350,4){\small{Back}}
\put(402,4){\small{Albedo}}
\put(454,4){\small{Shading}}
\caption{\small{\textbf{Additional qualitative examples from our method}.
}
\label{fig:qualitative_results2_supp}}
\end{figure*}

\begin{figure*}
\includegraphics[width=\linewidth]{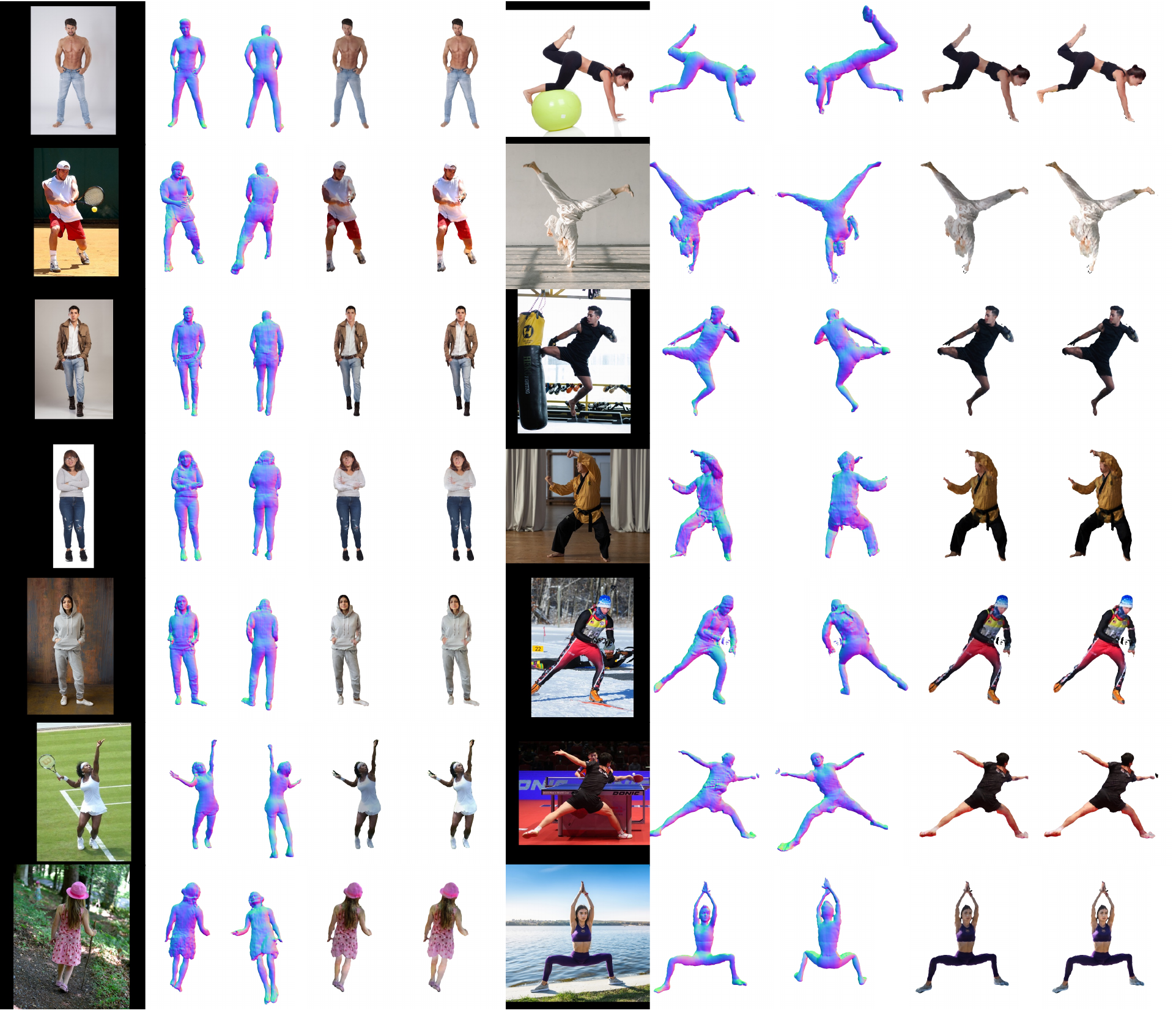}\\
\put(8,4){\small{Input image}}
\put(69,4){\small{Front}}
\put(102,4){\small{Back}}
\put(131,4){\small{Albedo}}
\put(170,4){\small{Shading}}
\put(223,4){\small{Input image}}
\put(288,4){\small{Front}}
\put(342,4){\small{Back}}
\put(394,4){\small{Albedo}}
\put(450,4){\small{Shading}}
\caption{\small{\textbf{Additional qualitative examples from our method}.
}
\label{fig:qualitative_results3_supp}}
\end{figure*}

\begin{figure*}
\includegraphics[width=\linewidth]{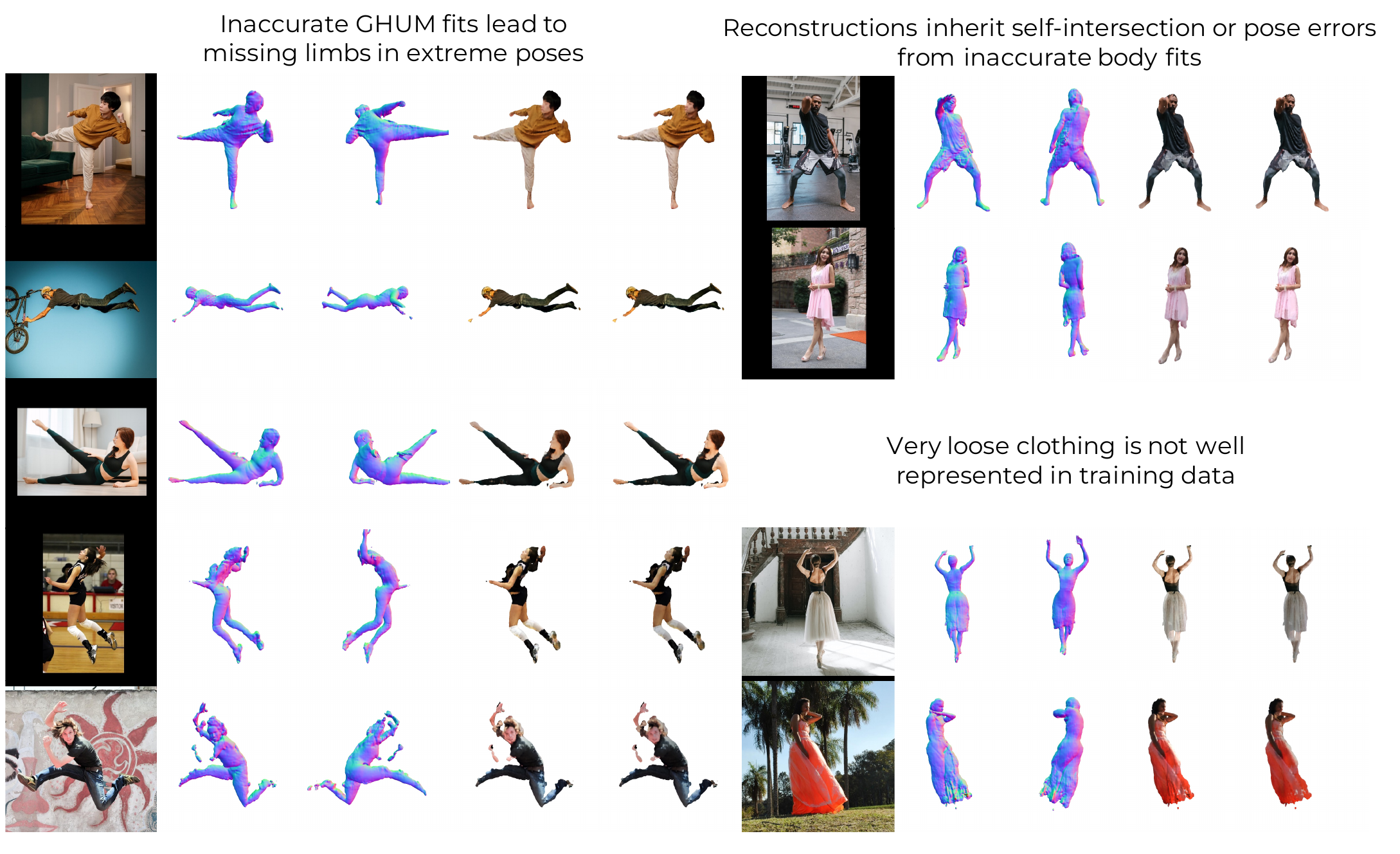}\\
\put(7,4){\small{Input image}}
\put(60,4){\small{Front geom.}}
\put(112,4){\small{Back geom.}}
\put(170,4){\small{Albedo}}
\put(220,4){\small{Shading}}
\put(265,4){\small{Input image}}
\put(312.5,4){\small{Front geom.}}
\put(363.5,4){\small{Back geom.}}
\put(409.5,4){\small{Albedo}}
\put(445.5,4){\small{Shading}}
\caption{\small{\textbf{Failure cases}.
We categorize failure cases in three main classes, all of them presumably due to noisy GHUM fits or insufficient training data. 
}
\label{fig:failure_cases_supp}}
\end{figure*}

\begin{figure*}
\includegraphics[width=0.95\linewidth]{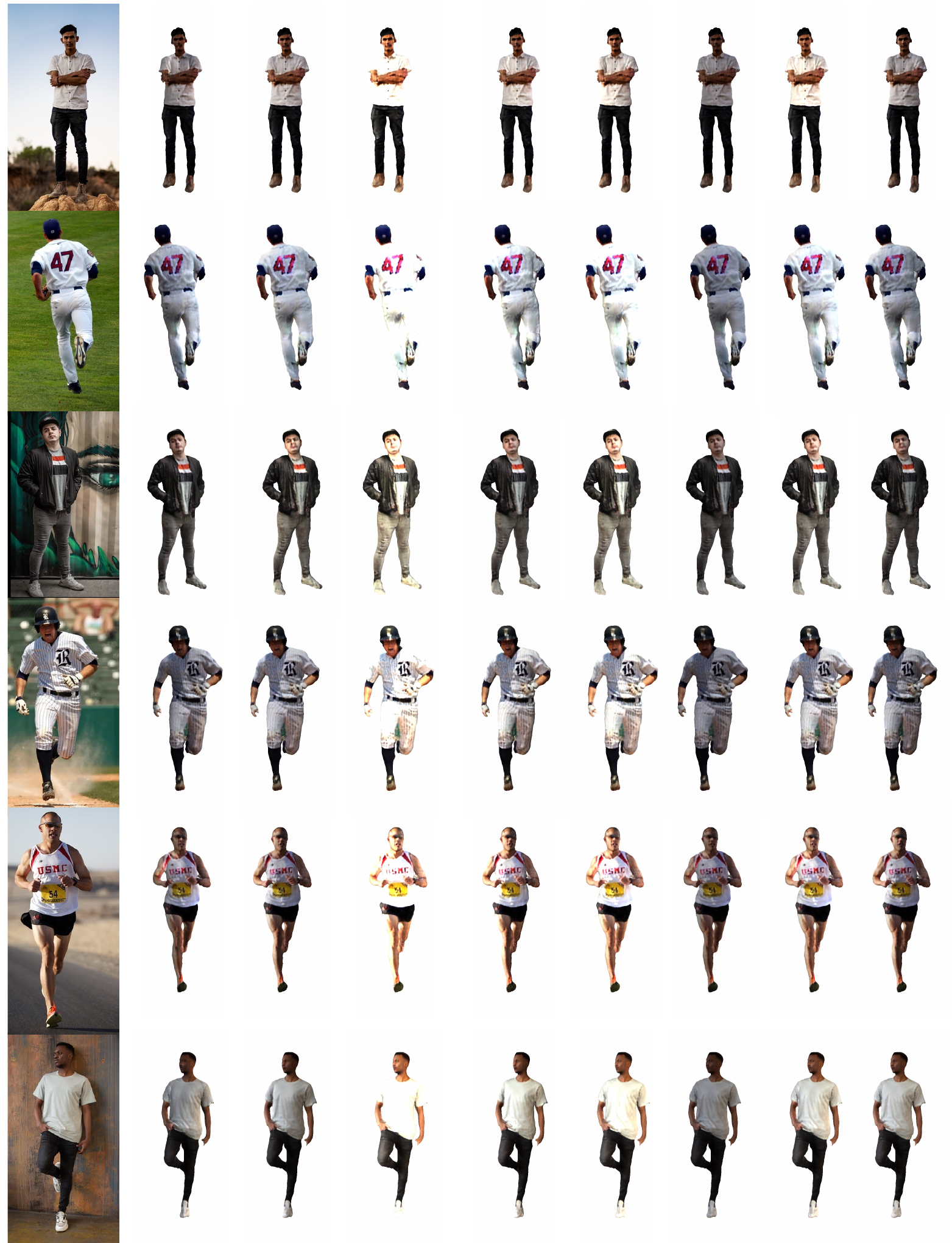}\\
\put(11,4){\small{Input image}}
\put(220,4){Relighted Reconstructions}
\caption{\small{\textbf{Qualitative results on 3D Human Relighting}.
}
\label{fig:shading_supp}}
\end{figure*}

\begin{figure*}
\includegraphics[width=\linewidth]{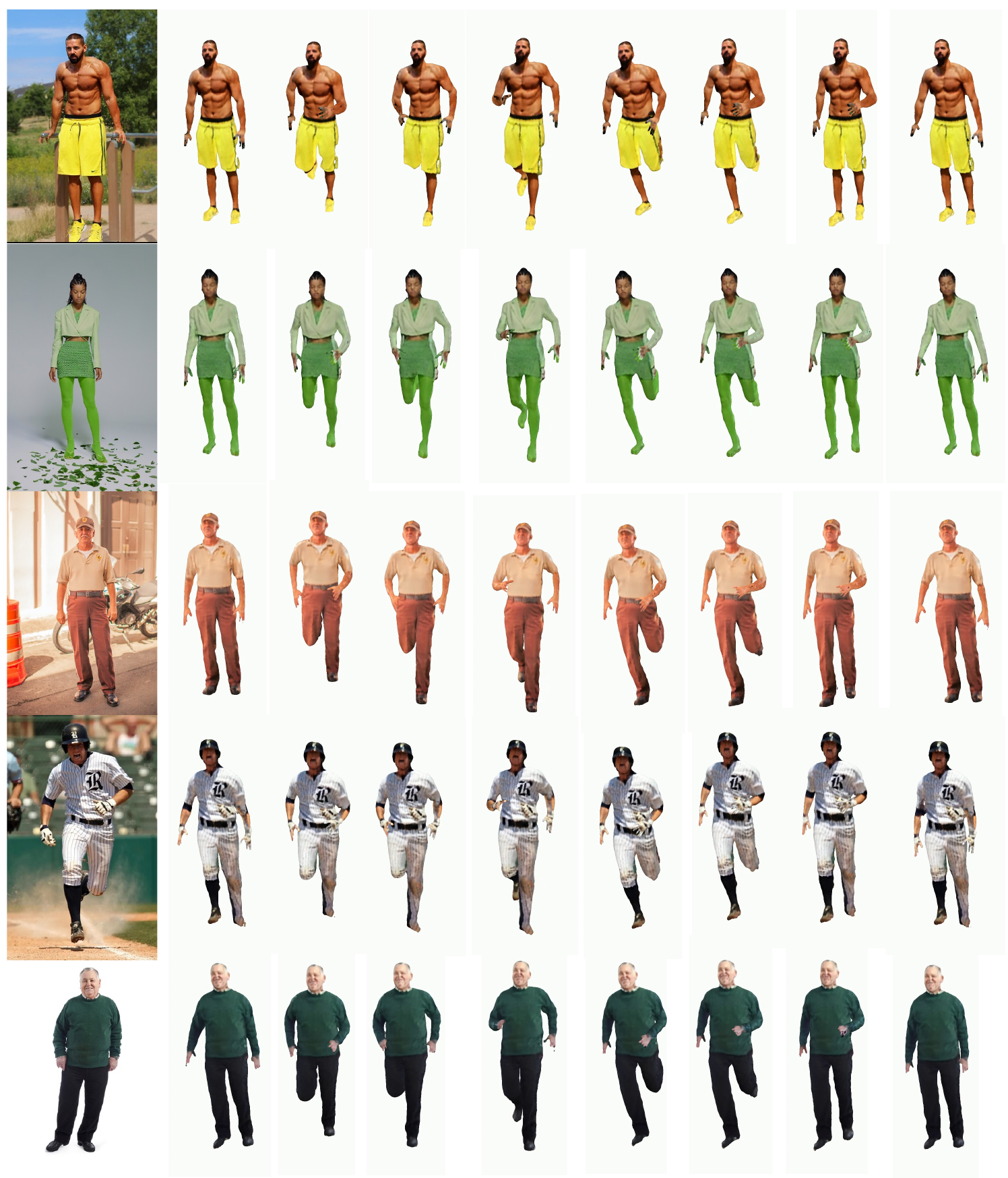}\\
\put(14,4){Input image}
\put(240,4){Animated Reconstructions}
\caption{\small{\textbf{Qualitative results on animation of 3D reconstructions}.
}
\label{fig:posing_supp}}
\end{figure*}

\begin{figure*}
\includegraphics[width=\linewidth]{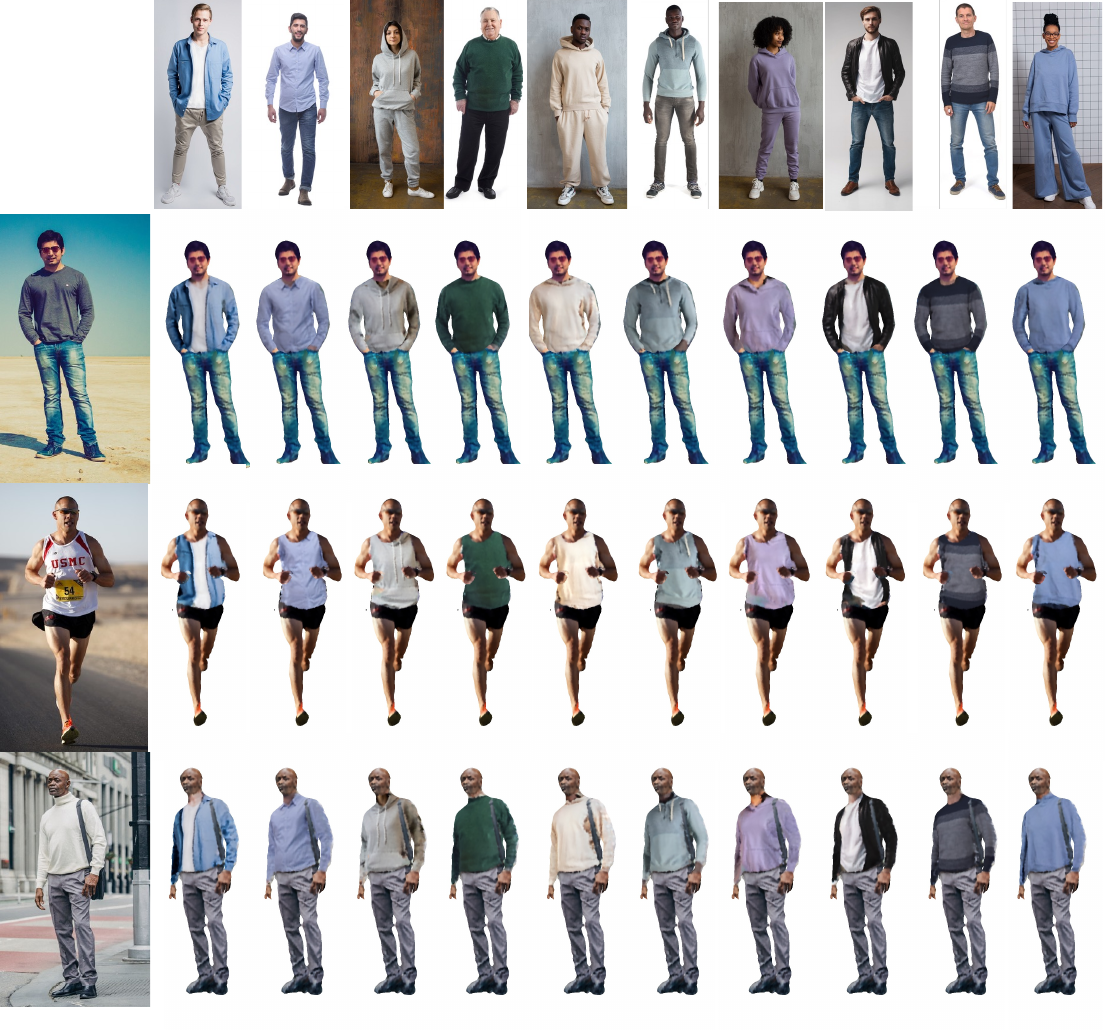}
\put(-250,470){Source clothing}
\put(-484,0){Input image}
\caption{\small{\textbf{More examples of cloth texture transfer}.
We extend the experiment from the main paper, showcasing an example of upper-body cloth try-on, and show the input images (left) and source clothing (upper-row). The 3D reconstructions look realistic and consistent accross all examples. Note that we only take one single image from both subject and clothing, and the network re-poses the affected S3Fs, allucinates occluded texture, and shades the clothing in the new scene.
}
\label{fig:editing_upper_supp}}
\end{figure*}

\begin{figure*}
\includegraphics[width=\linewidth]{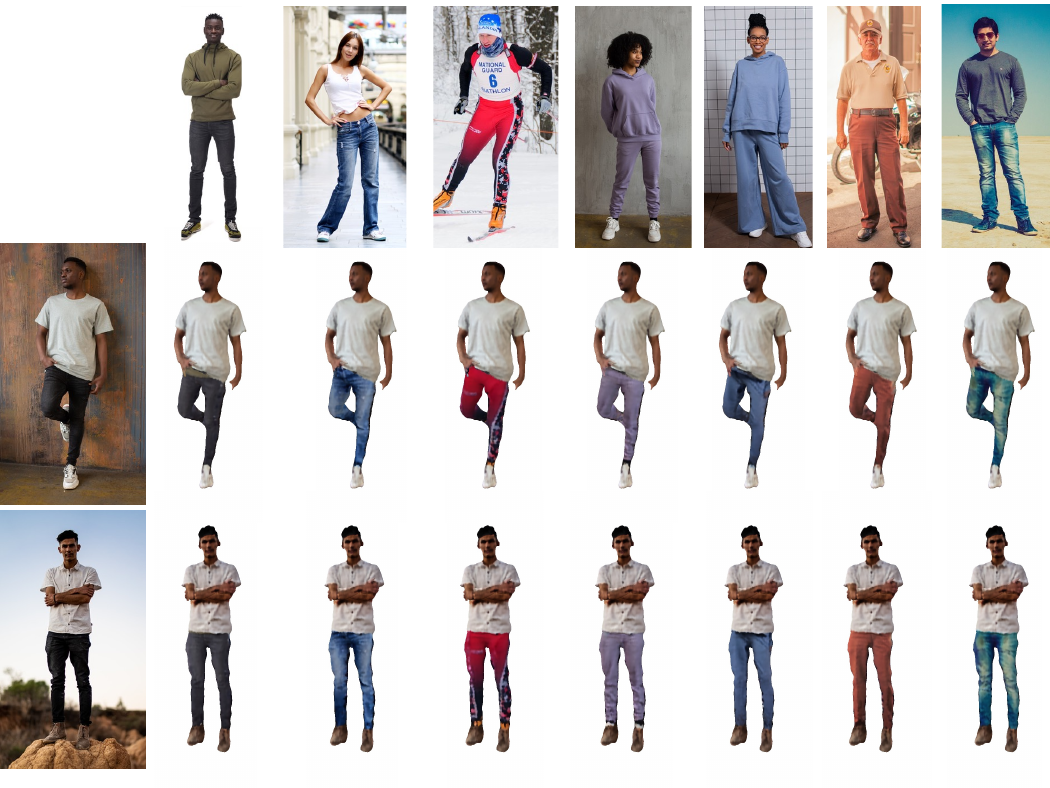}
\put(-220,375){Source clothing}
\put(-484,0){Input image}
\caption{\small{\textbf{More examples of cloth texture transfer}.
This example features try-on examples from lower-body clothing. See the Supplementary Video for more examples.
}
\label{fig:editing_lower_supp}}
\end{figure*}

Next, we extend the results presented in the main paper and provide more qualitative results on different tasks. We show more monocular 3D Human Reconstruction results in Figures~\ref{fig:qualitative_results1_supp},~\ref{fig:qualitative_results2_supp} and~\ref{fig:qualitative_results3_supp}, showcasing a variety of input poses, backgrounds, viewpoints and clothing.
We also categorize and present failure cases in Figure~\ref{fig:failure_cases_supp}, presumably all due to inaccurate GHUM fits or limited training data. We expect that better GHUM fits and accurate segmentations would lead to more consistent results on the presented images. 
Loose clothing is not well represented in our training set, since the dataset of real images used for training~\cite{bazavan2021hspace} comprises challenging poses of diverse sports mostly on tight clothing.
We also show additional results of 3D Human relighting and re-posing in Figures~\ref{fig:shading_supp} and~\ref{fig:posing_supp}, and extend the 3D virtual try-on experiments from the paper in Figures~\ref{fig:editing_upper_supp} and~\ref{fig:editing_lower_supp} with examples of upper-body and lower-body clothes respectively. The input images are shown in the left column and reference clothing is shown in the upper-row. Note that the transferred cloth textures are shaded consistently with the illumination from the original scenes, leading to photorealistic results that are coherent with the original non-edited 3D reconstruction. More examples are shown in the Supplementary Video, where we additionally show interpolations between cloth texture, body poses (animation) and scene illumination.
For cloth try-on, the interpolations showcase a remarkable level of robustness even when interpolating cloth features with very different textures, obtained from diverse body poses or shapes.

\begin{figure*}
\includegraphics[width=\linewidth]{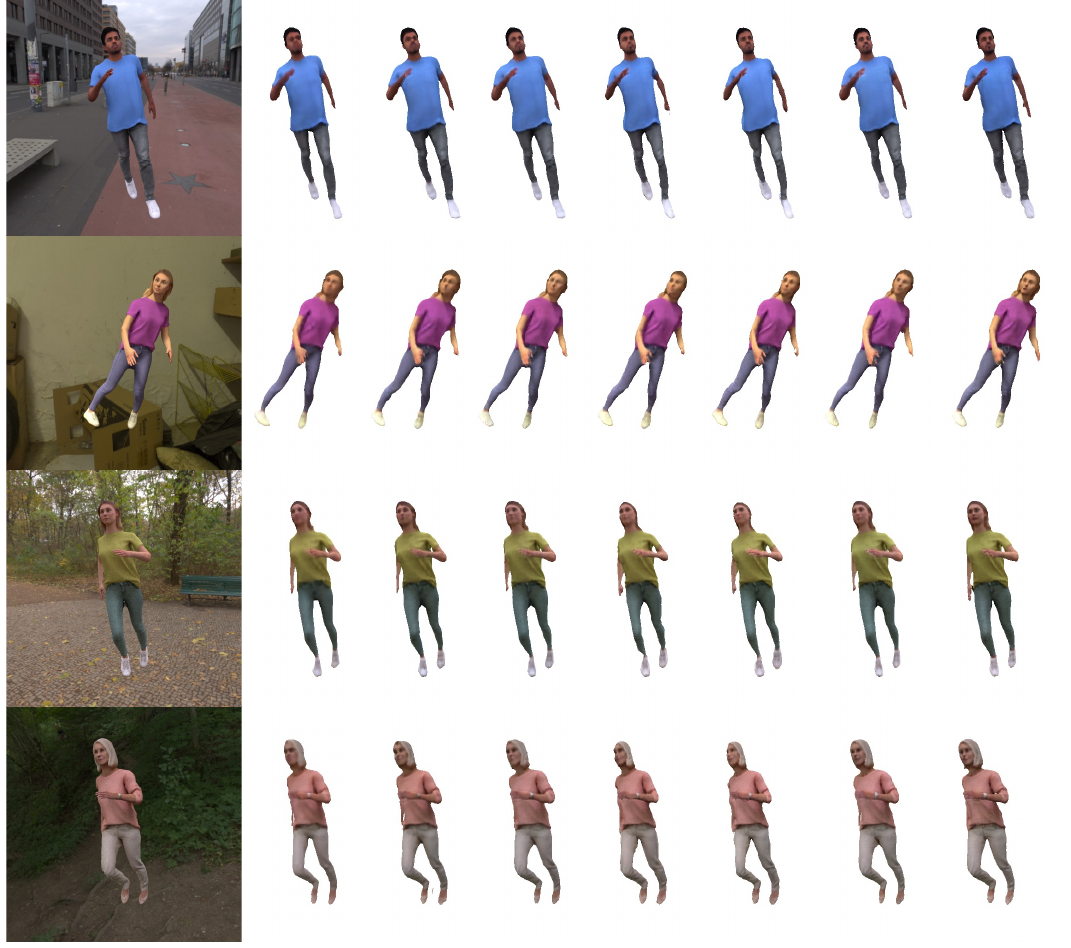}\\
\put(37,0){\small{Input image}}
\put(138,0){\small{2000}}
\put(190,0){\small{5000}}
\put(243,0){\small{8000}}
\put(288,0){\small{10000}}
\put(343,0){\small{12000}}
\put(397,0){\small{15000}}
\put(450,0){\small{18000}}
\put(200,-15){\small{Number of sampled points in the body surface}}
\caption{\small{\textbf{Qualitative ablation of number of points sampled in the body surface}, storing 3D Features. Our final model has 18000 points which provides the best tradeoff between GPU memory and sharpness. Lower number of points do not affect significantly quantitative results but are less capable of representing high-frequency details.
}
\label{fig:num_points_supp}}
\end{figure*}

Finally, we provide an intuition of the effect of the number of points/features in the results, in Figure~\ref{fig:num_points_supp}. Increasing the number of points leads to sharper reconstructions, although it does not significantly affect metrics quantitatively. Our final model has 18K points sampled on the body surface, which shows a good balance between memory consumption during training and texture sharpness.
\clearpage



\clearpage
{\small
\bibliographystyle{ieee_fullname}
\bibliography{references.bib}

\begin{thebibliography}{10}\itemsep=-1pt

\bibitem{alldieck2019learning}
Thiemo Alldieck, Marcus Magnor, Bharat~Lal Bhatnagar, Christian Theobalt, and
  Gerard Pons-Moll.
\newblock Learning to reconstruct people in clothing from a single rgb camera.
\newblock In {\em CVPR}, 2019.

\bibitem{alldieck2018detailed}
Thiemo Alldieck, Marcus Magnor, Weipeng Xu, Christian Theobalt, and Gerard
  Pons-Moll.
\newblock Detailed human avatars from monocular video.
\newblock In {\em 3DV}, 2018.

\bibitem{alldieck2018video}
Thiemo Alldieck, Marcus Magnor, Weipeng Xu, Christian Theobalt, and Gerard
  Pons-Moll.
\newblock Video based reconstruction of 3d people models.
\newblock In {\em CVPR}, 2018.

\bibitem{tex2shape}
Thiemo Alldieck, Gerard Pons-Moll, Christian Theobalt, and Marcus Magnor.
\newblock Tex2shape: Detailed full human body geometry from a single image.
\newblock In {\em ICCV}, 2019.

\bibitem{alldieck2020imghum}
Thiemo Alldieck, Hongyi Xu, and Cristian Sminchisescu.
\newblock {imGHUM}: Implicit generative models of {3D} human shape and
  articulated pose.
\newblock In {\em ICCV}, 2021.

\bibitem{phorhum}
Thiemo Alldieck, Mihai Zanfir, and Cristian Sminchisescu.
\newblock Photorealistic monocular 3d reconstruction of humans wearing
  clothing.
\newblock In {\em CVPR}, 2022.

\bibitem{bazavan2021hspace}
Eduard~Gabriel Bazavan, Andrei Zanfir, Mihai Zanfir, William~T Freeman, Rahul
  Sukthankar, and Cristian Sminchisescu.
\newblock Hspace: Synthetic parametric humans animated in complex environments.
\newblock {\em arXiv}, 2021.

\bibitem{bhatnagar2019multi}
Bharat~Lal Bhatnagar, Garvita Tiwari, Christian Theobalt, and Gerard Pons-Moll.
\newblock Multi-garment net: Learning to dress 3d people from images.
\newblock In {\em ICCV}, 2019.

\bibitem{blender}
{Blender Online Community}.
\newblock {\em Blender - a 3D modelling and rendering package}.
\newblock Blender Foundation, Blender Institute, Amsterdam, 2020.

\bibitem{cao22jiff}
Yukang Cao, Guanying Chen, Kai Han, Wenqi Yang, and Kwan-Yee~K. Wong.
\newblock Jiff: Jointly-aligned implicit face function for high quality single
  view clothed human reconstruction.
\newblock In {\em CVPR}, 2022.

\bibitem{chen2021animatable}
Jianchuan Chen, Ying Zhang, Di Kang, Xuefei Zhe, Linchao Bao, Xu Jia, and
  Huchuan Lu.
\newblock Animatable neural radiance fields from monocular rgb videos, 2021.

\bibitem{vgg_loss}
Qifeng Chen and Vladlen Koltun.
\newblock Photographic image synthesis with cascaded refinement networks.
\newblock In {\em ICCV}, 2017.

\bibitem{chen2019learning}
Zhiqin Chen and Hao Zhang.
\newblock Learning implicit fields for generative shape modeling.
\newblock In {\em CVPR}, 2019.

\bibitem{cmu_mocap}
Cmu graphics lab motion capture database.
\newblock \url{http://mocap.cs.cmu.edu/}.

\bibitem{corona2022lisa}
Enric Corona, Tomas Hodan, Minh Vo, Francesc Moreno-Noguer, Chris Sweeney,
  Richard Newcombe, and Lingni Ma.
\newblock Lisa: Learning implicit shape and appearance of hands.
\newblock {\em CVPR}, 2022.

\bibitem{corona2022learned}
Enric Corona, Gerard Pons-Moll, Guillem Aleny{\`a}, and Francesc Moreno-Noguer.
\newblock Learned vertex descent: A new direction for 3d human model fitting.
\newblock {\em ECCV}, 2022.

\bibitem{smplicit}
Enric Corona, Albert Pumarola, Guillem Alenya, Gerard Pons-Moll, and Francesc
  Moreno-Noguer.
\newblock Smplicit: Topology-aware generative model for clothed people.
\newblock In {\em CVPR}, 2021.

\bibitem{gabeur2019moulding}
Valentin Gabeur, Jean-S{\'e}bastien Franco, Xavier Martin, Cordelia Schmid, and
  Gregory Rogez.
\newblock Moulding humans: Non-parametric 3d human shape estimation from single
  images.
\newblock In {\em ICCV}, 2019.

\bibitem{glorot}
Xavier Glorot and Yoshua Bengio.
\newblock Understanding the difficulty of training deep feedforward neural
  networks.
\newblock In {\em Proceedings of the thirteenth international conference on
  artificial intelligence and statistics}, pages 249--256. JMLR Workshop and
  Conference Proceedings, 2010.

\bibitem{goodfellow2020generative}
Ian Goodfellow, Jean Pouget-Abadie, Mehdi Mirza, Bing Xu, David Warde-Farley,
  Sherjil Ozair, Aaron Courville, and Yoshua Bengio.
\newblock Generative adversarial networks.
\newblock {\em Communications of the ACM}, 63(11):139--144, 2020.

\bibitem{grishchenko2022blazepose}
Ivan Grishchenko, Valentin Bazarevsky, Andrei Zanfir, Eduard~Gabriel Bazavan,
  Mihai Zanfir, Richard Yee, Karthik Raveendran, Matsvei Zhdanovich, Matthias
  Grundmann, and Cristian Sminchisescu.
\newblock Blazepose ghum holistic: Real-time 3d human landmarks and pose
  estimation.
\newblock {\em arXiv}, 2022.

\bibitem{eikonal_loss}
Amos Gropp, Lior Yariv, Niv Haim, Matan Atzmon, and Yaron Lipman.
\newblock Implicit geometric regularization for learning shapes.
\newblock {\em ICML}, 2020.

\bibitem{hdri}
\url{https://polyhaven.com/}.

\bibitem{he2020geo}
Tong He, John Collomosse, Hailin Jin, and Stefano Soatto.
\newblock Geo-pifu: Geometry and pixel aligned implicit functions for
  single-view human reconstruction.
\newblock {\em NeurIPS}, 2020.

\bibitem{he2021arch++}
Tong He, Yuanlu Xu, Shunsuke Saito, Stefano Soatto, and Tony Tung.
\newblock Arch++: Animation-ready clothed human reconstruction revisited.
\newblock In {\em CVPR}, 2021.

\bibitem{huang2020arch}
Zeng Huang, Yuanlu Xu, Christoph Lassner, Hao Li, and Tony Tung.
\newblock Arch: Animatable reconstruction of clothed humans.
\newblock In {\em CVPR}, 2020.

\bibitem{jaegle2021perceiver}
Andrew Jaegle, Felix Gimeno, Andy Brock, Oriol Vinyals, Andrew Zisserman, and
  Joao Carreira.
\newblock Perceiver: General perception with iterative attention.
\newblock In {\em International conference on machine learning}, pages
  4651--4664. PMLR, 2021.

\bibitem{ji2022geometry}
Chaonan Ji, Tao Yu, Kaiwen Guo, Jingxin Liu, and Yebin Liu.
\newblock Geometry-aware single-image full-body human relighting.
\newblock In {\em ECCV}, 2022.

\bibitem{jiang2022selfrecon}
Boyi Jiang, Yang Hong, Hujun Bao, and Juyong Zhang.
\newblock Selfrecon: Self reconstruction your digital avatar from monocular
  video.
\newblock In {\em CVPR}, 2022.

\bibitem{jiang2022neuman}
Wei Jiang, Kwang~Moo Yi, Golnoosh Samei, Oncel Tuzel, and Anurag Ranjan.
\newblock Neuman: Neural human radiance field from a single video.
\newblock {\em ECCV}, 2022.

\bibitem{kanamori2019relighting}
Yoshihiro Kanamori and Yuki Endo.
\newblock Relighting humans: occlusion-aware inverse rendering for full-body
  human images.
\newblock {\em SIGGRAPH Asia}, 2019.

\bibitem{hmr}
Angjoo Kanazawa, Michael~J Black, David~W Jacobs, and Jitendra Malik.
\newblock End-to-end recovery of human shape and pose.
\newblock In {\em CVPR}, 2018.

\bibitem{adam}
Diederik~P Kingma and Jimmy Ba.
\newblock Adam: A method for stochastic optimization.
\newblock {\em arXiv}, 2014.

\bibitem{spin}
Nikos Kolotouros, Georgios Pavlakos, Michael~J Black, and Kostas Daniilidis.
\newblock Learning to reconstruct 3d human pose and shape via model-fitting in
  the loop.
\newblock In {\em ICCV}, 2019.

\bibitem{kolotouros2019convolutional}
Nikos Kolotouros, Georgios Pavlakos, and Kostas Daniilidis.
\newblock Convolutional mesh regression for single-image human shape
  reconstruction.
\newblock In {\em CVPR}, 2019.

\bibitem{lagunas2021single}
Manuel Lagunas, Xin Sun, Jimei Yang, Ruben Villegas, Jianming Zhang, Zhixin
  Shu, Belen Masia, and Diego Gutierrez.
\newblock Single-image full-body human relighting.
\newblock {\em ECCV}, 2022.

\bibitem{lassner2017unite}
Christoph Lassner, Javier Romero, Martin Kiefel, Federica Bogo, Michael~J
  Black, and Peter~V Gehler.
\newblock Unite the people: Closing the loop between 3d and 2d human
  representations.
\newblock In {\em CVPR}, 2017.

\bibitem{li2020self}
Peike Li, Yunqiu Xu, Yunchao Wei, and Yi Yang.
\newblock Self-correction for human parsing.
\newblock {\em PAMI}, 2020.

\bibitem{li2021topologically}
Tianye Li, Shichen Liu, Timo Bolkart, Jiayi Liu, Hao Li, and Yajie Zhao.
\newblock Topologically consistent multi-view face inference using volumetric
  sampling.
\newblock In {\em ICCV}, 2021.

\bibitem{smpl}
Matthew Loper, Naureen Mahmood, Javier Romero, Gerard Pons-Moll, and Michael~J
  Black.
\newblock Smpl: A skinned multi-person linear model.
\newblock {\em ToG}, 2015.

\bibitem{marching_cubes}
William~E Lorensen and Harvey~E Cline.
\newblock Marching cubes: A high resolution 3d surface construction algorithm.
\newblock {\em SIGGRAPH}, 1987.

\bibitem{mescheder2018training}
Lars Mescheder, Andreas Geiger, and Sebastian Nowozin.
\newblock Which training methods for gans do actually converge?
\newblock In {\em International conference on machine learning}, pages
  3481--3490. PMLR, 2018.

\bibitem{mescheder2019occupancy}
Lars Mescheder, Michael Oechsle, Michael Niemeyer, Sebastian Nowozin, and
  Andreas Geiger.
\newblock Occupancy networks: Learning 3d reconstruction in function space.
\newblock In {\em CVPR}, 2019.

\bibitem{keypointnerf}
Marko Mihajlovic, Aayush Bansal, Michael Zollhoefer, Siyu Tang, and Shunsuke
  Saito.
\newblock {KeypointNeRF}: Generalizing image-based volumetric avatars using
  relative spatial encoding of keypoints.
\newblock In {\em ECCV}, 2022.

\bibitem{nerf}
Ben Mildenhall, Pratul~P Srinivasan, Matthew Tancik, Jonathan~T Barron, Ravi
  Ramamoorthi, and Ren Ng.
\newblock Nerf: Representing scenes as neural radiance fields for view
  synthesis.
\newblock {\em ECCV}, 2020.

\bibitem{mir20pix2surf}
Aymen Mir, Thiemo Alldieck, and Gerard Pons-Moll.
\newblock Learning to transfer texture from clothing images to 3d humans.
\newblock In {\em CVPR}, 2020.

\bibitem{natsume2019siclope}
Ryota Natsume, Shunsuke Saito, Zeng Huang, Weikai Chen, Chongyang Ma, Hao Li,
  and Shigeo Morishima.
\newblock Siclope: Silhouette-based clothed people.
\newblock In {\em CVPR}, 2019.

\bibitem{omran2018neural}
Mohamed Omran, Christoph Lassner, Gerard Pons-Moll, Peter Gehler, and Bernt
  Schiele.
\newblock Neural body fitting: Unifying deep learning and model based human
  pose and shape estimation.
\newblock In {\em 3DV}, 2018.

\bibitem{onizuka2020tetratsdf}
Hayato Onizuka, Zehra Hayirci, Diego Thomas, Akihiro Sugimoto, Hideaki
  Uchiyama, and Rin-ichiro Taniguchi.
\newblock Tetratsdf: 3d human reconstruction from a single image with a
  tetrahedral outer shell.
\newblock In {\em CVPR}, 2020.

\bibitem{deepsdf}
Jeong~Joon Park, Peter Florence, Julian Straub, Richard Newcombe, and Steven
  Lovegrove.
\newblock Deepsdf: Learning continuous signed distance functions for shape
  representation.
\newblock In {\em CVPR}, 2019.

\bibitem{smplx}
Georgios Pavlakos, Vasileios Choutas, Nima Ghorbani, Timo Bolkart, Ahmed~AA
  Osman, Dimitrios Tzionas, and Michael~J Black.
\newblock Expressive body capture: 3d hands, face, and body from a single
  image.
\newblock In {\em CVPR}, 2019.

\bibitem{peng2021animatable}
Sida Peng, Junting Dong, Qianqian Wang, Shangzhan Zhang, Qing Shuai, Xiaowei
  Zhou, and Hujun Bao.
\newblock Animatable neural radiance fields for modeling dynamic human bodies.
\newblock In {\em CVPR}, 2021.

\bibitem{neuralbody}
Sida Peng, Yuanqing Zhang, Yinghao Xu, Qianqian Wang, Qing Shuai, Hujun Bao,
  and Xiaowei Zhou.
\newblock Neural body: Implicit neural representations with structured latent
  codes for novel view synthesis of dynamic humans.
\newblock In {\em CVPR}, 2021.

\bibitem{pumarola20193dpeople}
Albert Pumarola, Jordi Sanchez-Riera, Gary Choi, Alberto Sanfeliu, and Francesc
  Moreno-Noguer.
\newblock 3dpeople: Modeling the geometry of dressed humans.
\newblock In {\em ICCV}, 2019.

\bibitem{swish}
Prajit Ramachandran, Barret Zoph, and Quoc~V Le.
\newblock Searching for activation functions.
\newblock {\em arXiv preprint arXiv:1710.05941}, 2017.

\bibitem{remelli2022drivable}
Edoardo Remelli, Timur Bagautdinov, Shunsuke Saito, Chenglei Wu, Tomas Simon,
  Shih-En Wei, Kaiwen Guo, Zhe Cao, Fabian Prada, Jason Saragih, et~al.
\newblock Drivable volumetric avatars using texel-aligned features.
\newblock In {\em SIGGRAPH}, 2022.

\bibitem{renderpeople}
Renderpeople dataset.
\newblock \url{https://renderpeople.com/}.

\bibitem{mano}
Javier Romero, Dimitrios Tzionas, and Michael~J Black.
\newblock Embodied hands: Modeling and capturing hands and bodies together.
\newblock {\em ToG}, 2017.

\bibitem{frankmocap}
Yu Rong, Takaaki Shiratori, and Hanbyul Joo.
\newblock Frankmocap: Fast monocular 3d hand and body motion capture by
  regression and integration.
\newblock {\em arXiv}, 2020.

\bibitem{unet}
Olaf Ronneberger, Philipp Fischer, and Thomas Brox.
\newblock U-net: Convolutional networks for biomedical image segmentation.
\newblock In {\em International Conference on Medical image computing and
  computer-assisted intervention}, pages 234--241. Springer, 2015.

\bibitem{saito2019pifu}
Shunsuke Saito, Zeng Huang, Ryota Natsume, Shigeo Morishima, Angjoo Kanazawa,
  and Hao Li.
\newblock Pifu: Pixel-aligned implicit function for high-resolution clothed
  human digitization.
\newblock In {\em ICCV}, 2019.

\bibitem{saito2020pifuhd}
Shunsuke Saito, Tomas Simon, Jason Saragih, and Hanbyul Joo.
\newblock Pifuhd: Multi-level pixel-aligned implicit function for
  high-resolution 3d human digitization.
\newblock In {\em CVPR}, 2020.

\bibitem{smith2019facsimile}
David Smith, Matthew Loper, Xiaochen Hu, Paris Mavroidis, and Javier Romero.
\newblock Facsimile: Fast and accurate scans from an image in less than a
  second.
\newblock In {\em ICCV}, 2019.

\bibitem{su2021nerf}
Shih-Yang Su, Frank Yu, Michael Zollhoefer, and Helge Rhodin.
\newblock A-nerf: Surface-free human 3d pose refinement via neural rendering.
\newblock {\em NeurIPS}, 2021.

\bibitem{tajima2021relighting}
Daichi Tajima, Yoshihiro Kanamori, and Yuki Endo.
\newblock Relighting humans in the wild: Monocular full-body human relighting
  with domain adaptation.
\newblock In {\em CGF}, 2021.

\bibitem{tewari2022advances}
Ayush Tewari, Justus Thies, Ben Mildenhall, Pratul Srinivasan, Edgar Tretschk,
  W Yifan, Christoph Lassner, Vincent Sitzmann, Ricardo Martin-Brualla, Stephen
  Lombardi, et~al.
\newblock Advances in neural rendering.
\newblock In {\em CGF}, 2022.

\bibitem{varol2018bodynet}
Gul Varol, Duygu Ceylan, Bryan Russell, Jimei Yang, Ersin Yumer, Ivan Laptev,
  and Cordelia Schmid.
\newblock Bodynet: Volumetric inference of 3d human body shapes.
\newblock In {\em ECCV}, 2018.

\bibitem{vaswani2017attention}
Ashish Vaswani, Noam Shazeer, Niki Parmar, Jakob Uszkoreit, Llion Jones,
  Aidan~N Gomez, {\L}ukasz Kaiser, and Illia Polosukhin.
\newblock Attention is all you need.
\newblock {\em NeurIPS}, 2017.

\bibitem{arah_eccv22}
Shaofei Wang, Katja Schwarz, Andreas Geiger, and Siyu Tang.
\newblock Arah: Animatable volume rendering of articulated human sdfs.
\newblock In {\em ECCV}, 2022.

\bibitem{weng2022humannerf}
Chung-Yi Weng, Brian Curless, Pratul~P Srinivasan, Jonathan~T Barron, and Ira
  Kemelmacher-Shlizerman.
\newblock Humannerf: Free-viewpoint rendering of moving people from monocular
  video.
\newblock In {\em CVPR}, 2022.

\bibitem{xiu2022icon}
Yuliang Xiu, Jinlong Yang, Dimitrios Tzionas, and Michael~J Black.
\newblock Icon: Implicit clothed humans obtained from normals.
\newblock In {\em CVPR}, 2022.

\bibitem{xu2021h}
Hongyi Xu, Thiemo Alldieck, and Cristian Sminchisescu.
\newblock H-nerf: Neural radiance fields for rendering and temporal
  reconstruction of humans in motion.
\newblock {\em NeurIPS}, 2021.

\bibitem{xu2020ghum}
Hongyi Xu, Eduard~Gabriel Bazavan, Andrei Zanfir, William~T Freeman, Rahul
  Sukthankar, and Cristian Sminchisescu.
\newblock Ghum \& ghuml: Generative 3d human shape and articulated pose models.
\newblock In {\em CVPR}, 2020.

\bibitem{xu20213d}
Xiangyu Xu and Chen~Change Loy.
\newblock 3d human texture estimation from a single image with transformers.
\newblock In {\em CVPR}, 2021.

\bibitem{yang2021s3}
Ze Yang, Shenlong Wang, Sivabalan Manivasagam, Zeng Huang, Wei-Chiu Ma, Xinchen
  Yan, Ersin Yumer, and Raquel Urtasun.
\newblock S3: Neural shape, skeleton, and skinning fields for 3d human
  modeling.
\newblock In {\em CVPR}, 2021.

\bibitem{integratedpifu}
Kennard Yanting~Chan, Guosheng Lin, Haiyu Zhao, and Weisi Lin.
\newblock Integratedpifu: Integrated pixel aligned implicit function for
  single-view human reconstruction.
\newblock In {\em ECCV}, 2022.

\bibitem{volsdf}
Lior Yariv, Jiatao Gu, Yoni Kasten, and Yaron Lipman.
\newblock Volume rendering of neural implicit surfaces.
\newblock {\em NeurIPS}, 2021.

\bibitem{zanfir2020weakly}
Andrei Zanfir, Eduard~Gabriel Bazavan, Hongyi Xu, William~T Freeman, Rahul
  Sukthankar, and Cristian Sminchisescu.
\newblock Weakly supervised 3d human pose and shape reconstruction with
  normalizing flows.
\newblock In {\em ECCV}, pages 465--481. Springer, 2020.

\bibitem{zanfir2021thundr}
Mihai Zanfir, Andrei Zanfir, Eduard~Gabriel Bazavan, William~T Freeman, Rahul
  Sukthankar, and Cristian Sminchisescu.
\newblock Thundr: Transformer-based 3d human reconstruction with markers.
\newblock In {\em CVPR}, 2021.

\bibitem{implicit_hessian_loss}
Jingyang Zhang, Yao Yao, Shiwei Li, Tian Fang, David McKinnon, Yanghai Tsin,
  and Long Quan.
\newblock Critical regularizations for neural surface reconstruction in the
  wild.
\newblock In {\em Proceedings of the IEEE/CVF Conference on Computer Vision and
  Pattern Recognition}, pages 6270--6279, 2022.

\bibitem{zheng2022avatar}
Yufeng Zheng, Victoria~Fern{\'a}ndez Abrevaya, Marcel~C B{\"u}hler, Xu Chen,
  Michael~J Black, and Otmar Hilliges.
\newblock Im avatar: Implicit morphable head avatars from videos.
\newblock In {\em CVPR}, 2022.

\bibitem{zheng2021pamir}
Zerong Zheng, Tao Yu, Yebin Liu, and Qionghai Dai.
\newblock Pamir: Parametric model-conditioned implicit representation for
  image-based human reconstruction.
\newblock {\em PAMI}, 2021.

\bibitem{zheng2019deephuman}
Zerong Zheng, Tao Yu, Yixuan Wei, Qionghai Dai, and Yebin Liu.
\newblock Deephuman: 3d human reconstruction from a single image.
\newblock In {\em ICCV}, 2019.

\bibitem{zhu2019detailed}
Hao Zhu, Xinxin Zuo, Sen Wang, Xun Cao, and Ruigang Yang.
\newblock Detailed human shape estimation from a single image by hierarchical
  mesh deformation.
\newblock In {\em CVPR}, 2019.

\end{thebibliography}
}

\end{document}